\documentclass[preprints,article,accept,moreauthors,pdftex]{Definitions/mdpi}

\usepackage{rotating} 
\usepackage{multirow}
\usepackage{booktabs}
\usepackage{tabularx} 
\usepackage{algorithm}
\usepackage{algorithmic}
\usepackage{amsmath} 
\usepackage{amsmath}  
\usepackage{graphicx}

\usepackage{algorithm}

\usepackage[utf8]{inputenc}
\usepackage{textgreek}
\usepackage{booktabs}
\usepackage{graphicx} 
\usepackage{array}

\usepackage{booktabs}
\usepackage{graphicx}  
\usepackage{array}
\usepackage{pifont} 
\usepackage{placeins} 
\usepackage{float}    
\usepackage{amssymb}  
\usepackage{caption} 

\usepackage{xcolor}
\firstpage{1} 
\pubvolume{1}
\issuenum{1}
\articlenumber{0}
\pubyear{2022}
\copyrightyear{2022}
\datereceived{} 
\dateaccepted{} 
\datepublished{} 
\hreflink{https://doi.org/} 

\Title{GAFR-Net: A Graph Attention and Fuzzy-Rule Network for Interpretable Breast Cancer Image Classification}

\TitleCitation{Title}

\TitleCitation{ Lin-Guo Gao
}

\Title{GAFR-Net: A Graph Attention and Fuzzy-Rule Network for Interpretable Breast Cancer Image Classification}

\TitleCitation{Title}

\Author{Lin-Guo Gao $^{1,2}$ and Suxing Liu $^{1,2,*}$}

\AuthorNames{Lin-Guo Gao and Suxing Liu}

\AuthorCitation{Gao, L.-G.; Liu, S.}

\address{%
$^{1}$ \quad Department of IT Engineering, Mokwon University, Daejeon 35349, South Korea; \\
$^{2}$ \quad School of Digital Arts, Jiangxi Arts \& Ceramics Technology Institute, Jingdezhen 333001, China
}

\corres{Correspondence: bentondoucet@gmail.com}
\abstract{Accurate classification of breast cancer histopathology images is pivotal for early oncological diagnosis and therapeutic intervention. However, conventional deep learning architectures often encounter performance degradation under limited annotations and suffer from a "black-box" nature, hindering their clinical integration. To mitigate these limitations, we propose GAFR-Net, a robust and interpretable Graph Attention and Fuzzy-Rule Network specifically engineered for histopathology image classification with scarce supervision. GAFR-Net constructs a similarity-driven graph representation to model inter-sample relationships and employs a multi-head graph attention mechanism to capture complex relational features across heterogeneous tissue structures. Concurrently, a differentiable fuzzy-rule module encodes intrinsic topological descriptors—including node degree, clustering coefficient, and label consistency—into explicit, human-understandable diagnostic logic. This design establishes transparent "IF-THEN" mappings that mimic the heuristic deduction process of medical experts, providing clear reasoning behind each prediction without relying on post-hoc attribution methods. Extensive evaluations on three benchmark datasets (BreakHis, Mini-DDSM, and ICIAR2018) demonstrate that GAFR-Net consistently outperforms various state-of-the-art methods across multiple magnifications and classification tasks. These results validate the superior generalization and practical utility of GAFR-Net as a reliable decision-support tool for weakly supervised medical image analysis.}


\begin{document}

\section{Introduction}

Breast cancer remains one of the most prevalent malignant tumors among women worldwide, and accurate early diagnosis is essential for improving patient survival rates \cite{b1,b2}. Conventional histopathological examination relies heavily on the subjective assessment of experienced pathologists, which may lead to inter-observer variability, increased diagnostic workload, and a non-negligible risk of oversight or misinterpretation \cite{b3}. In recent years, deep learning techniques—particularly Convolutional Neural Networks (CNNs)—have demonstrated remarkable success in automated medical image classification \cite{b25,b26}. Despite these advances, CNN-based architectures are inherently constrained by their local receptive fields, limiting their ability to capture global relational dependencies among histopathological samples. Moreover, under small-sample or class-imbalanced conditions, CNN models are susceptible to feature bias, unstable optimization, and overfitting \cite{b27,b28}. More critically, their opaque “black-box” decision-making process lacks interpretability, which hinders clinical trust and restricts their adoption in real-world diagnostic workflows.

To overcome these limitations, this study proposes a novel interpretable graph-based learning framework termed GAFR-Net (Graph Attention and Fuzzy-Rule Network). The proposed model constructs a similarity-driven graph representation among histopathology images, enabling the learning process to jointly exploit sample-wise semantic relationships and global topological context. A multi-head Graph Attention Network (GAT) is employed to adaptively weight neighborhood contributions, thereby capturing relational features that are difficult for conventional CNNs to model. In addition, GAFR-Net extracts a set of topology-aware descriptors—including clustering coefficient, node degree, and two-hop label agreement—to characterize the structural role of each node within the graph. These descriptors not only provide complementary structural information but also serve as an explicit basis for interpretable reasoning. To further enhance transparency and robustness, we integrate a trainable fuzzy-rule reasoning module that modulates node representations through explicit “IF–THEN” rules derived from topological patterns. For instance, a representative rule can be formulated as: “IF the node degree is high AND the label consistency is strong, THEN the sample belongs to a high-confidence region.” Such rule-based reasoning establishes a direct and human-understandable link between graph topology and diagnostic predictions, facilitating trustworthy decision support in clinical applications.

Our contributions can be summarized as follows:

\begin{itemize}
\item We propose GAFR-Net, a novel interpretable classification framework that tightly integrates graph attention mechanisms with trainable fuzzy-rule reasoning, enabling unified relational learning and intrinsic model interpretability.

\item We design a topology-aware feature representation incorporating clustering coefficient, node degree, and two-hop label agreement, which effectively captures inter-sample structural relationships and enhances generalization under class-imbalanced settings.

\item We introduce a trainable fuzzy-rule module that establishes explicit “IF–THEN” mappings between topological descriptors and classification outputs, providing transparent and human-interpretable diagnostic reasoning.

\item Extensive experiments on three public datasets demonstrate that GAFR-Net achieves superior accuracy, robustness, and interpretability compared with CNN-based, Transformer-based, and graph-based baselines, highlighting its potential for real-world weakly supervised medical image analysis.
\end{itemize}

\begin{figure}[t]
    \centering
    \includegraphics[width=0.98\textwidth]{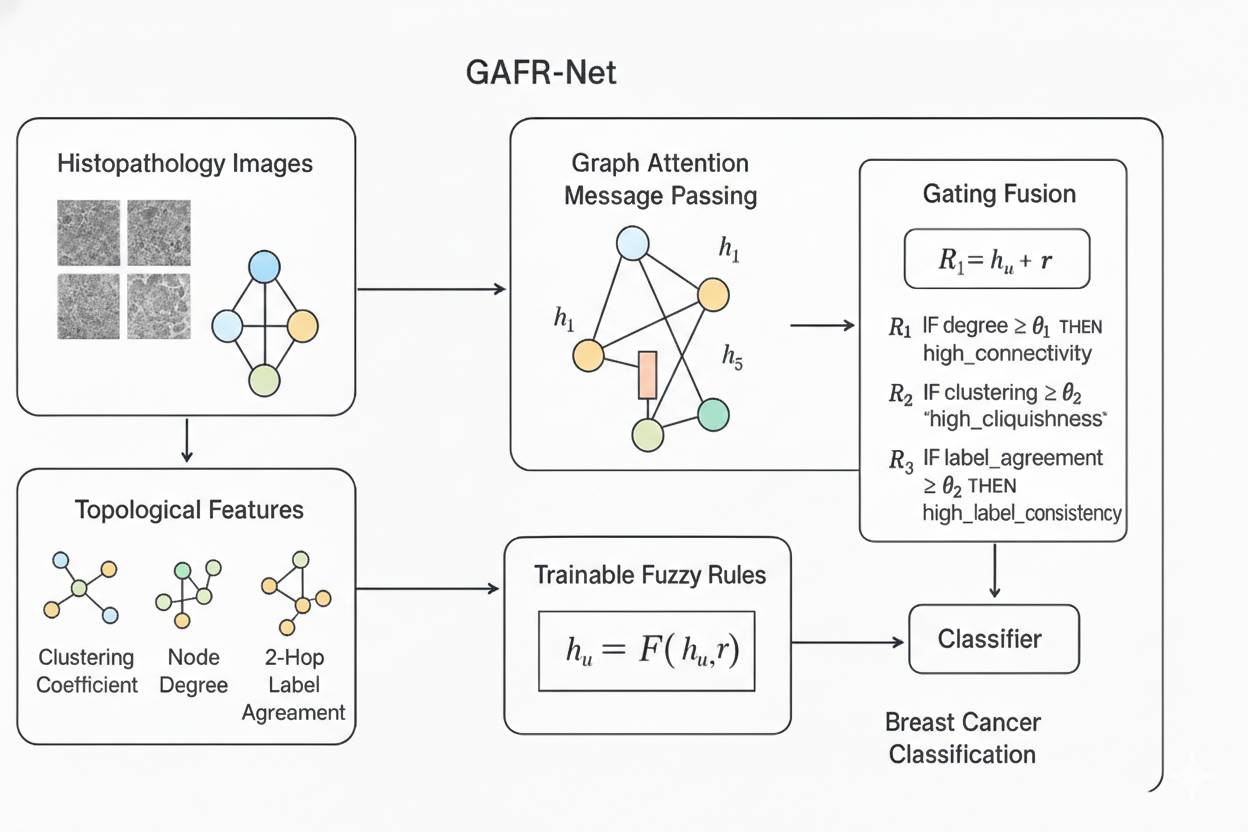}
    \caption{Overall pipeline of the proposed GAFR-Net, consisting of graph construction, 
    topological descriptor extraction, graph attention message passing, fuzzy-rule reasoning,
    and gating fusion for interpretable breast cancer histopathology classification.}
    \label{fig:gafr_framework}
\end{figure}

\section{Related Work}

\subsection{Evolution of Structural Representation: From CNNs to GNNs}
The paradigm of automated histopathology image classification has been long defined by the success of Convolutional Neural Networks (CNNs). Early deep learning frameworks focused on utilizing dense, grid-based architectures like ResNet and DenseNet to extract high-level semantic features from breast cancer tissue slides \cite{Zhou2025}. Despite achieving competitive accuracy, these methods possess inherent structural vulnerabilities. Standard convolutions rely on a fixed receptive field, which makes them inherently suboptimal for modeling the non-Euclidean, irregular spatial distributions of heterogeneous cell nuclei and stroma. Consequently, CNNs often fail to capture the long-range topological dependencies that are critical for characterizing the tumor microenvironment  \cite{Younesi2024,Chowdhury2023}.

To transcend these limitations, Graph Neural Networks (GNNs) have emerged as a powerful alternative by abstracting histopathology images into graph-structured data. By representing individual tissue patches or nuclei as nodes and their spatial interactions as edges, GNNs can perform relational reasoning via sophisticated message-passing mechanisms \cite{ArshadChoudhry2025}. Recent advancements have integrated Multi-head Graph Attention (GAT) to dynamically assign importance to neighboring nodes, thereby enhancing the robustness of feature aggregation against data noise \cite{Sun2023Attention, Zhang2023Graph}. However, a major gap persists: while these graph-based architectures improve representational power, they typically operate as "black boxes." In clinical settings, high performance alone is insufficient; the lack of transparency in how GNNs fuse relational features limits their accountability and clinical trustworthiness \cite{Alves2025AI}.
\subsection{Interpretable Reasoning and Knowledge-driven Diagnosis}
As AI systems transition into clinical practice, interpretability has become a paramount requirement for oncology diagnosis. Existing interpretability techniques mainly fall into post-hoc attribution methods, such as Grad-CAM or LIME, which generate visual heatmaps to highlight salient regions \cite{Abdullakutty2024}. Nevertheless, these visual cues often lack explicit logical grounding and fail to explain ``why'' a specific diagnostic decision was reached in a way that aligns with pathological guidelines. Unlike these post-hoc methods that require secondary models to interpret a ``black-box'' predictor, fuzzy logic and rule-based systems provide an intrinsic, self-interpretable solution. By offering a transparent ``IF-THEN'' reasoning mechanism that mimics the heuristic deduction process of medical experts \cite{Stefanelli2019}, fuzzy reasoning can handle the inherent uncertainty and imprecision in medical data by mapping quantitative descriptors into linguistic variables.

The current research landscape highlights two unresolved challenges: (1) conventional CNNs and Transformers struggle to adequately represent the complex global topological information of breast tissue under limited annotation conditions, and (2) state-of-the-art GNNs rarely incorporate human-understandable, symbolic reasoning pathways directly into their learning process \cite{Tiwari2025}. To bridge these critical gaps, we propose GAFR-Net, a unified framework that synergizes graph attention-based structural modeling with a trainable, self-interpretable fuzzy-rule reasoning layer. Specifically, GAFR-Net extracts discriminative topological cues such as node degree, clustering coefficients, and label consistency---and encodes them into a set of explicit diagnostic rules \cite{Li2025Pathological}. This integration not only enhances the model's ability to learn from scarce annotations through relational priors but also transforms the relational embeddings into transparent, rule-driven diagnostic logic . By embedding interpretability directly into the architecture, GAFR-Net ensures that each classification is accompanied by an accountable reasoning path, thereby supporting reliable clinical decision-making.
\begin{algorithm}[t]
\caption{GAFR-Net: Graph Attention and Fuzzy Rule Network}
\label{alg:gafr}
\small
\begin{algorithmic}[1]
\renewcommand{\algorithmicrequire}{\textbf{Input:}}
\renewcommand{\algorithmicensure}{\textbf{Output:}}

\REQUIRE Image set $\{x_i\}$, similarity threshold $\tau$
\ENSURE Predictions $\hat{y}$ and fuzzy rule activations

\STATE \textbf{Step 1: Graph Construction}
\FOR{ each pair $(x_i, x_j)$ }
    \IF{ $s_{ij} > \tau$ }
        \STATE Add edge $(i,j)$ to graph $G$
    \ENDIF
\ENDFOR

\STATE \textbf{Step 2: Topological Feature Extraction}
\FOR{ each node $u$ }
    \STATE Compute clustering $C(u)$, degree $d(u)$, label agreement $L(u)$
    \STATE $f_u \gets [C(u), d(u), L(u)]$
\ENDFOR

\STATE \textbf{Step 3: Graph Attention Message Passing}
\FOR{ each GAT layer $l$ }
    \STATE Compute attention $\alpha_{uv}$ for neighbors $v$
    \STATE $h_u^{l+1} \gets \sigma\left(\sum_{v \in \mathcal{N}(u)} \alpha_{uv} \mathbf{W} h_v^l\right)$
\ENDFOR

\STATE \textbf{Step 4: Trainable Fuzzy Rule Reasoning}
\FOR{ each node $u$ }
    \STATE $r_k(u) \gets \mu_k(f_u)$ for each rule $k$
    \STATE $r(u) \gets \sum_k \alpha_k r_k(u)$
\ENDFOR

\STATE \textbf{Step 5: Gating Fusion and Prediction}
\STATE $h'_u \gets h_u + r(u)$
\STATE $\hat{y}_u \gets \text{Softmax}(\mathbf{W}_c h'_u)$

\end{algorithmic}
\end{algorithm}

\section{Methodology}

In this section, we present the overall framework of GAFR-Net, an interpretable graph-based learning model designed for breast cancer histopathology classification. The proposed architecture integrates four key components: (1) similarity-based graph construction, (2) topology-aware feature extraction, (3) multi-head graph attention message passing, and (4) a trainable fuzzy-rule reasoning module. A final gating fusion layer combines relational embeddings with fuzzy-rule activations for robust and interpretable classification. The complete procedure is summarized in Algorithm 1.

\begin{figure}[t]
    \centering
    \includegraphics[width=1.0\linewidth]{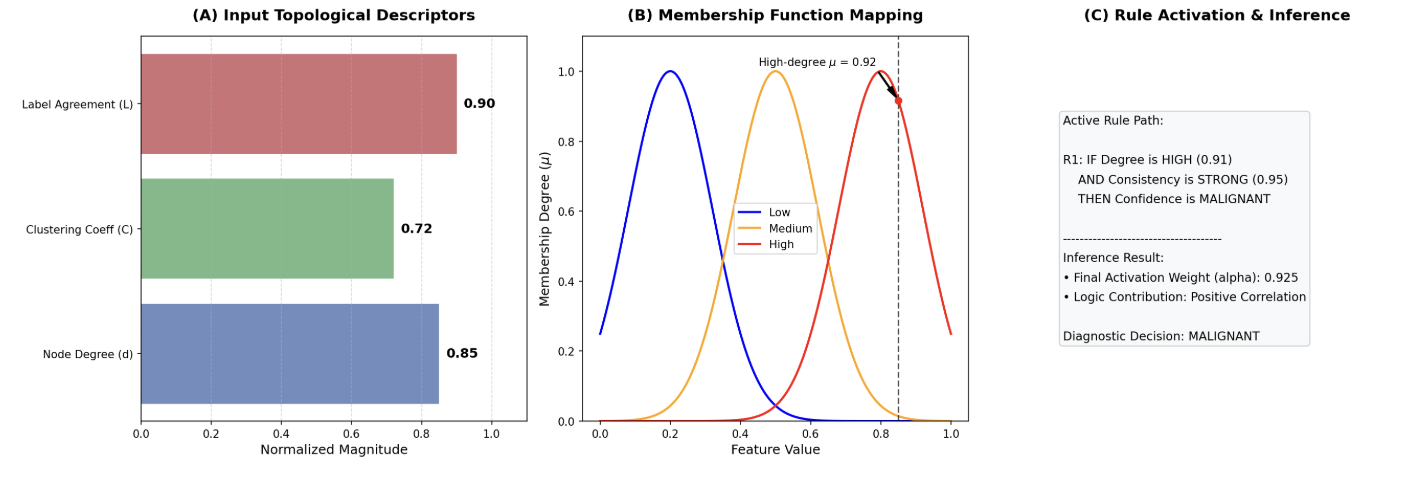}
    \caption{The intrinsic reasoning mechanism of GAFR-Net. (A) Input topological descriptors representing the local and global graph structure; (B) Feature mapping through trainable Gaussian membership functions to handle uncertainty; (C) Explicit ``IF-THEN'' logic activation that provides an accountable diagnostic path.}
    \label{fig:fuzzy_process}
\end{figure}

\begin{figure}[t]
    \centering
    \begin{minipage}{0.2\textwidth}
        \centering
        \includegraphics[width=0.8\textwidth, height=0.8\textwidth]{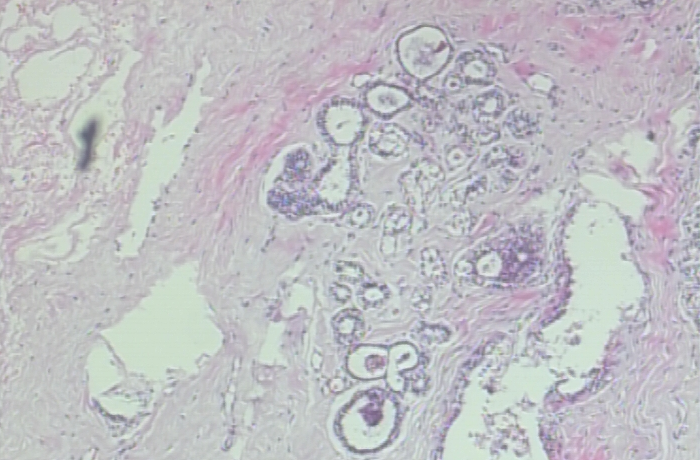}\\
        40x
    \end{minipage}
    \begin{minipage}{0.2\textwidth}
        \centering
        \includegraphics[width=0.8\textwidth, height=0.8\textwidth]{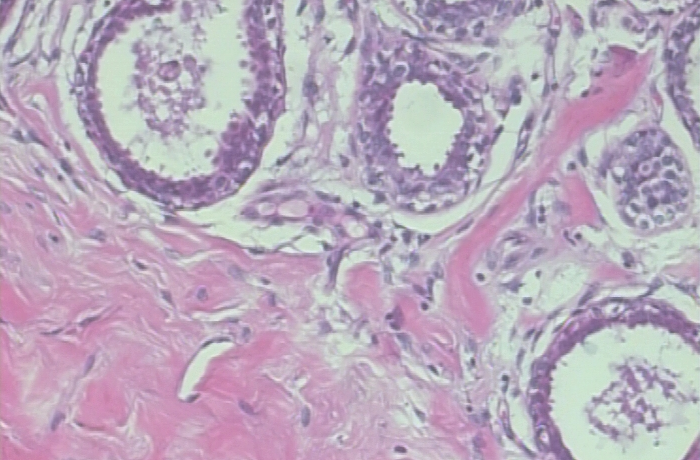}\\
        100x
    \end{minipage}
    
    \vspace{1em}
    
    \begin{minipage}{0.2\textwidth}
        \centering
        \includegraphics[width=0.8\textwidth, height=0.8\textwidth]{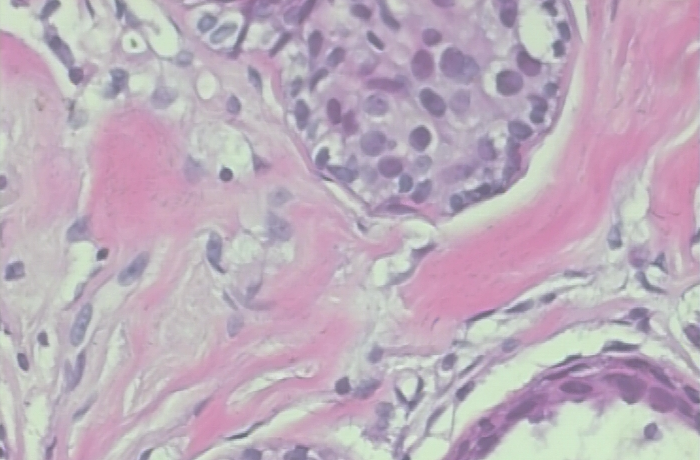}\\
        200x
    \end{minipage}
    \begin{minipage}{0.2\textwidth}
        \centering
        \includegraphics[width=0.8\textwidth, height=0.8\textwidth]{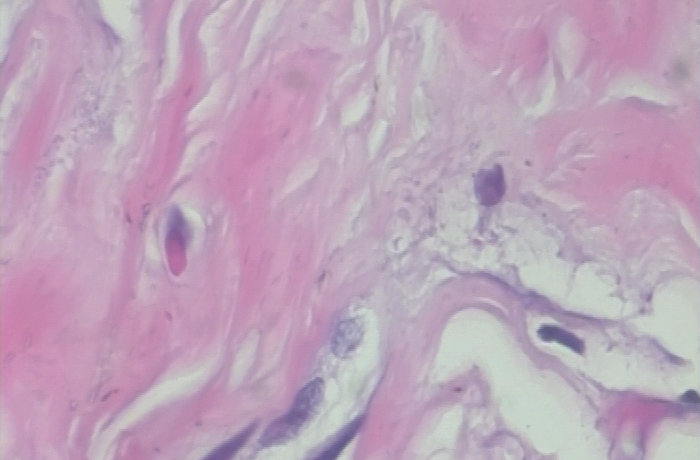}\\
        400x
    \end{minipage}
    
    \caption{Images of Breast Cancer Histopathology from the BreakHis Dataset at Four Magnifications}
    \label{Tal6}
\end{figure}

\begin{figure}[t]
    \centering
    \begin{minipage}{0.2\textwidth}
        \centering
        \includegraphics[width=0.8\textwidth, height=0.8\textwidth]{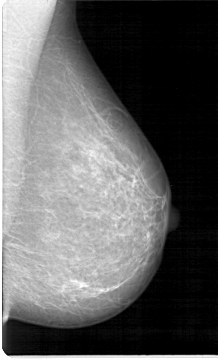}\\
        Normal 
    \end{minipage}
    \begin{minipage}{0.2\textwidth}
        \centering
        \includegraphics[width=0.8\textwidth, height=0.8\textwidth]{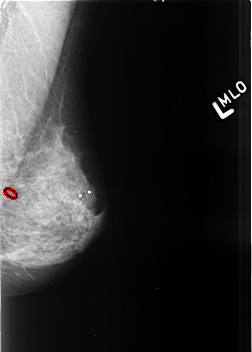}\\
        Benign
    \end{minipage}
    
    \vspace{1em}
    
            ·    \begin{minipage}{0.2\textwidth}
        \centering
        \includegraphics[width=0.8\textwidth, height=0.8\textwidth]{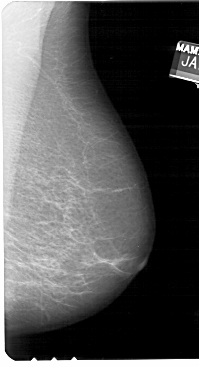}\\
        Malignant
    \end{minipage}

    \caption{Images of Breast Cancer Histopathology from the Mini-DDSM Dataset}
    \label{Tal6}
\end{figure}

\begin{figure}[t]
    \centering
    \begin{minipage}{0.2\textwidth}
        \centering
        \includegraphics[width=0.8\textwidth, height=0.8\textwidth]{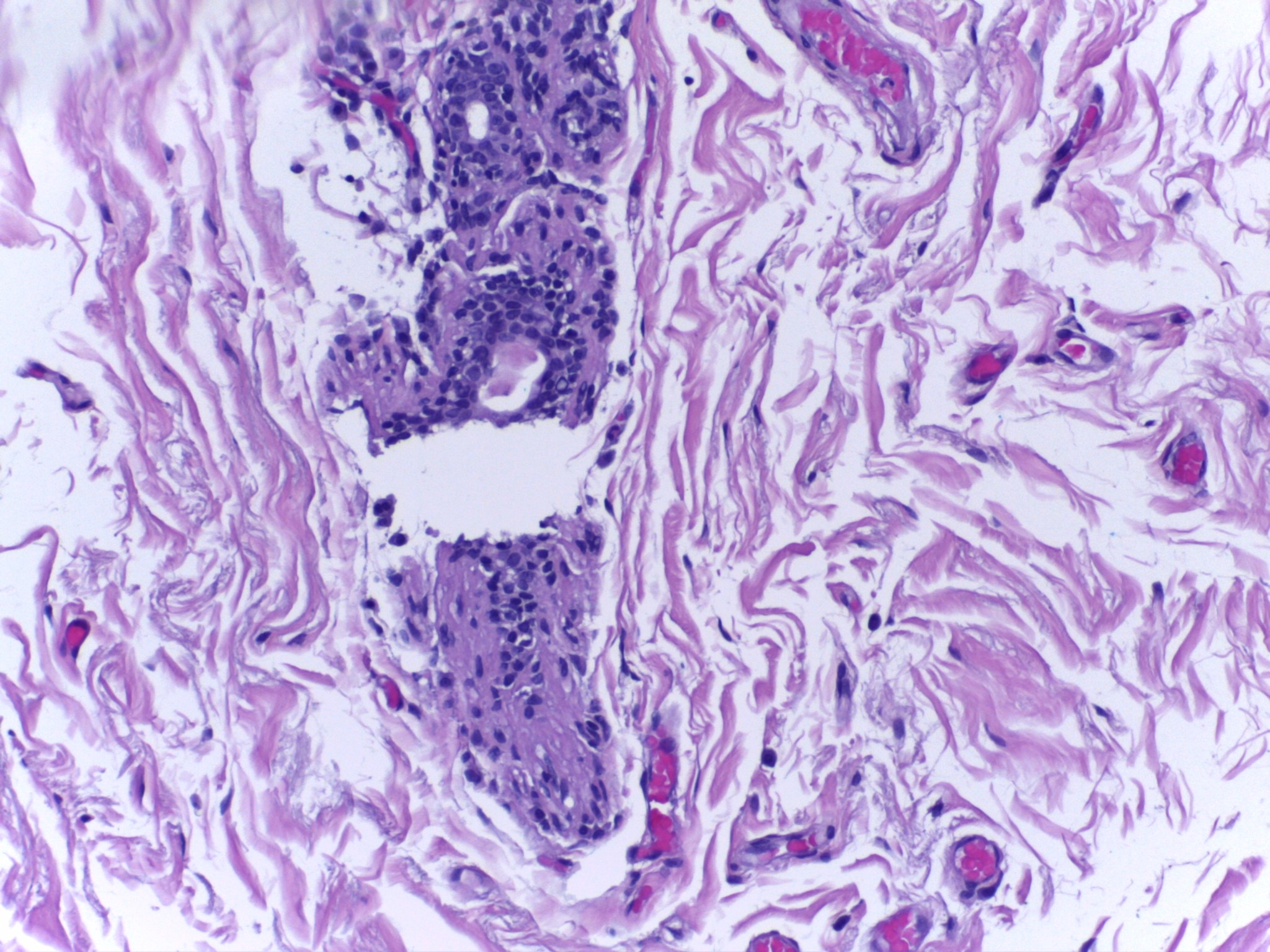}\\
        Normal
    \end{minipage}
    \begin{minipage}{0.2\textwidth}
        \centering
        \includegraphics[width=0.8\textwidth, height=0.8\textwidth]{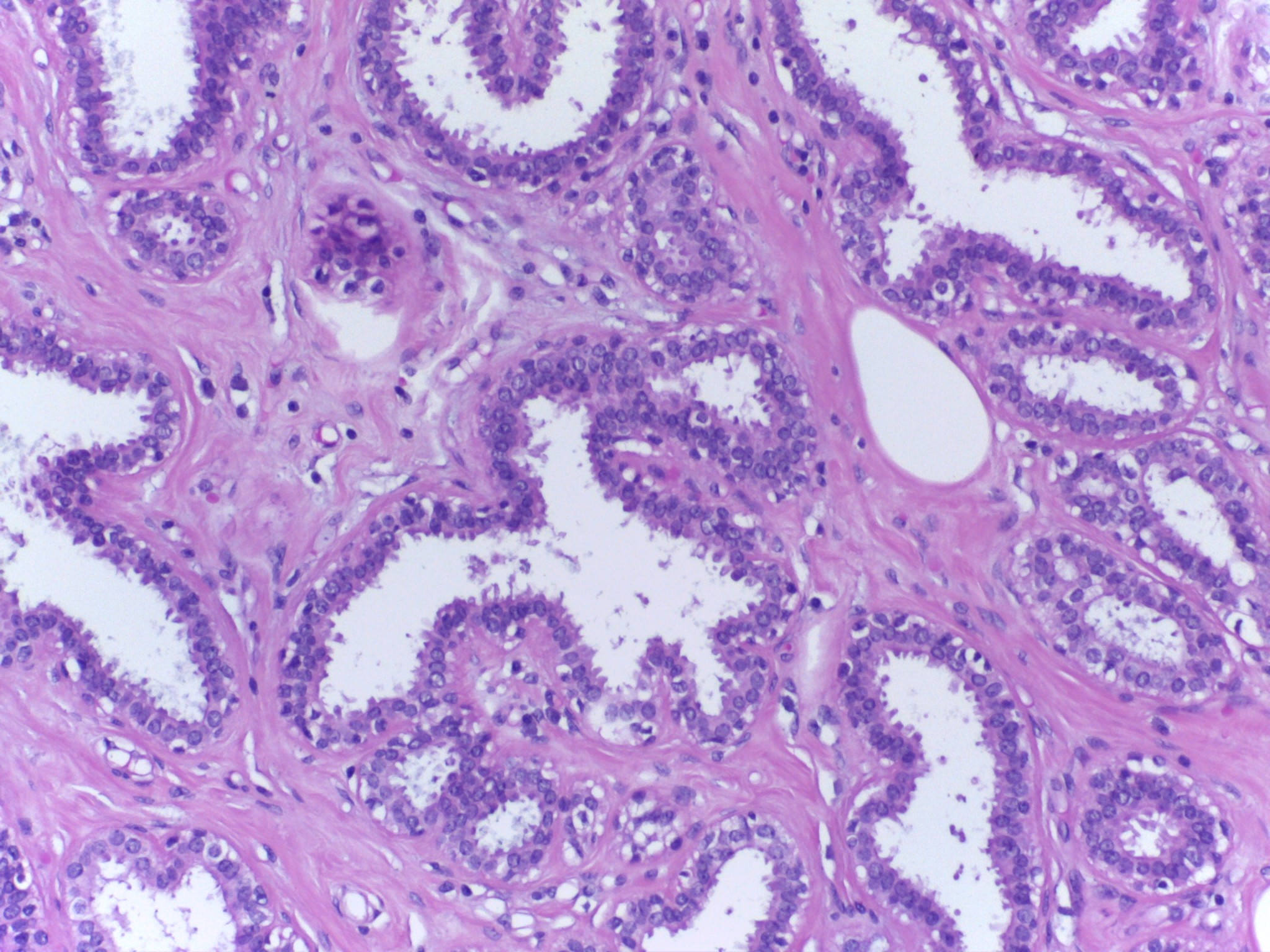}\\
        Benign
    \end{minipage}
    
    \vspace{1em}
    
    \begin{minipage}{0.2\textwidth}
        \centering
        \includegraphics[width=0.8\textwidth, height=0.8\textwidth]{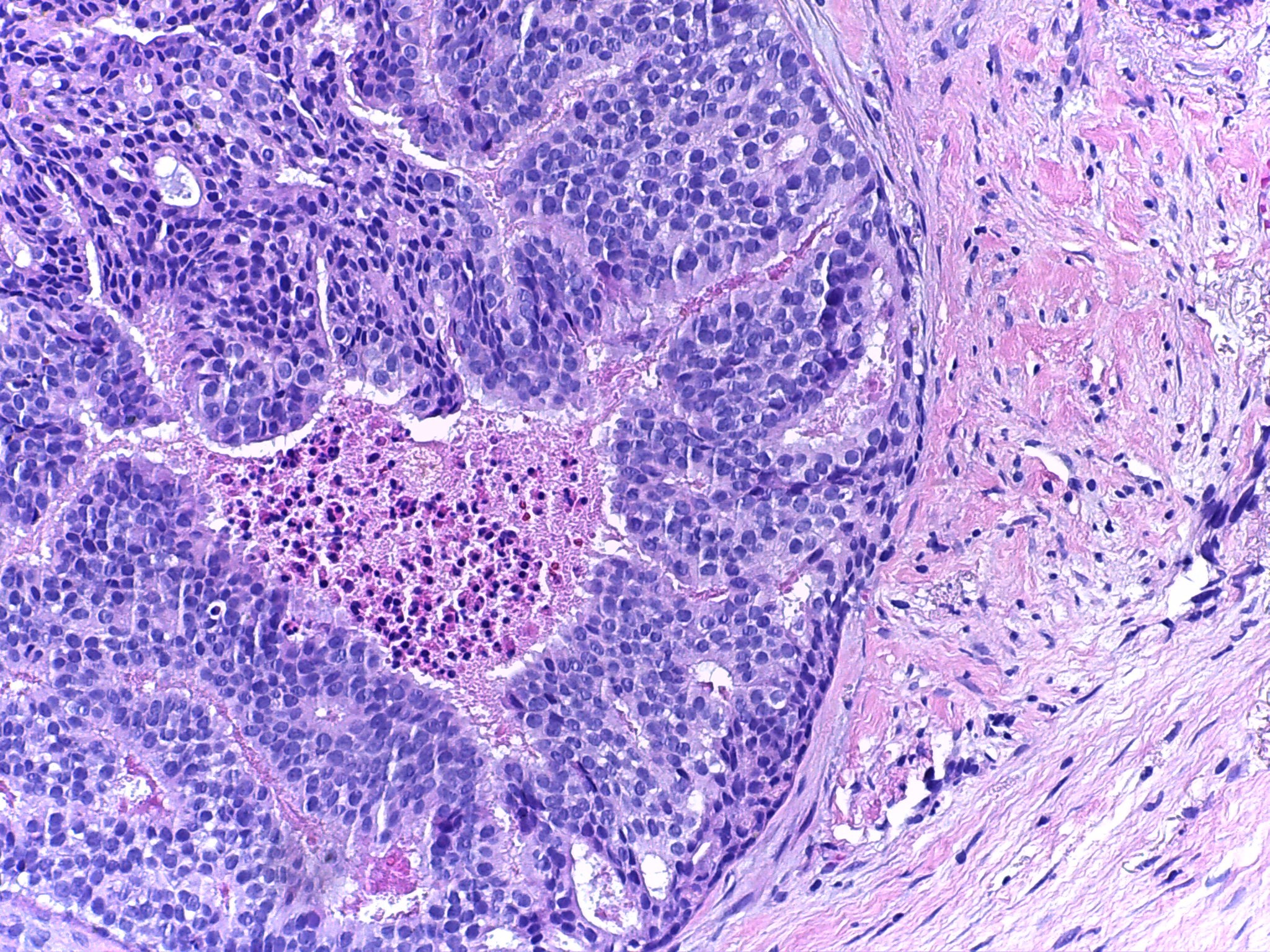}\\
        InSitu
    \end{minipage}
    \begin{minipage}{0.2\textwidth}
        \centering
        \includegraphics[width=0.8\textwidth, height=0.8\textwidth]{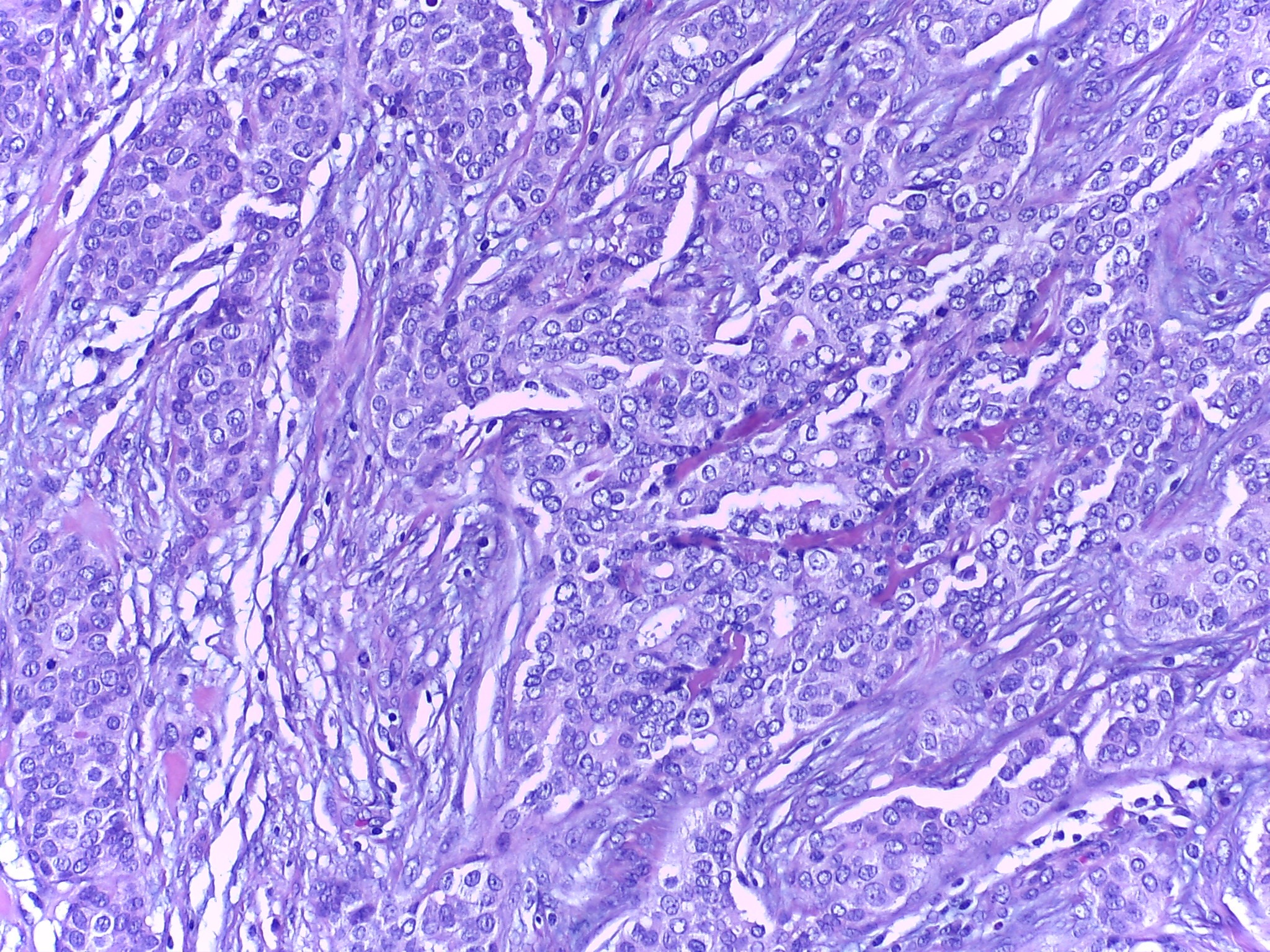}\\
        Invasive
    \end{minipage}

    \caption{Images of Breast Cancer Histopathology from the  ICIAR2018  Dataset }
    \label{Tal6}
\end{figure}

\subsection{Graph Construction }
Given a set of histopathology images $\{x_i\}_{i=1}^N$, we construct a graph $G=(V,E)$ where each node represents an image and edges encode semantic similarity\cite{Chan2023}. For each image pair $(x_i, x_j)$, a similarity score $s_{ij}$ is computed. An edge is added between nodes $i$ and $j$ if:
\begin{equation}
s_{ij} > \tau,
\end{equation}
where $\tau$ is a threshold controlling graph sparsity. This structure allows the model to capture inter-sample relationships often missed by conventional CNNs.

\subsection{Topological Feature Extraction }
To enhance interpretability, we compute three topological descriptors for each node $u$:
\begin{itemize}
    \item Clustering coefficient $C(u)$: indicates local neighborhood density.
    \item Node degree $d(u)$: represents graph 
    \item Two-hop label agreement $L(u)$: reflects class consistency among extended neighbors.
\end{itemize}
These form a topology vector $f_u = [C(u), d(u), L(u)]$, which serves as input to the fuzzy-rule module.

\subsection{Graph Attention Message Passing }
To capture heterogeneous neighborhood influence, we employ multi-head Graph Attention Network (GAT) layers\cite{Sun2023}. For each node $u$ and its neighbors $v \in \mathcal{N}(u)$, an attention coefficient $\alpha_{uv}$ weights the contribution of $v$:
\begin{equation}
h^{(l+1)}_u = \sigma \left( \sum_{v \in \mathcal{N}(u)} \alpha_{uv} \mathbf{W} h^{(l)}_v \right),
\end{equation}
where $\mathbf{W}$ is a learnable projection matrix and $\sigma$ is an activation function.

\subsection{Trainable Fuzzy Rule Reasoning}

To facilitate human-understandable explanations, we design a differentiable fuzzy-rule module that transforms abstract embeddings into symbolic logic\cite{Yin2024Hierarchical}. For each node $u$ and rule $k$, a membership degree is computed as follows:
\begin{equation}
r_k(u) = \mu_k(f_u),
\end{equation}
where $\mu_k(\cdot)$ denotes a trainable membership function that maps numerical topological descriptors into linguistic variables. The overall rule activation is then defined by:
\begin{equation}
r(u) = \sum_{k} \alpha_k r_k(u),
\end{equation}
where $\alpha_k$ represents the learnable importance weights assigned to each diagnostic rule. This module effectively maps graph topology into explicit reasoning logic, such as: \textit{"IF node degree is HIGH AND label agreement is STRONG, THEN the sample belongs to a HIGH-CONFIDENCE region"}.

The proposed fuzzy reasoning workflow is visualized in Figure 2. As illustrated, the module first normalizes the extracted topological descriptors, including node degree, clustering coefficient, and label consistency (Figure 4A)\cite{Cang2018Representability}. These numerical magnitudes are subsequently mapped into overlapping linguistic sets through differentiable Gaussian membership functions (Figure 4B)\cite{Pei2024Credibility}. Finally, the framework generates a transparent diagnostic logic path by aggregating these activations (Figure 4C)\cite{Muhammad2025ALL}. Unlike conventional ``black-box'' architectures or post-hoc attribution methods that provide secondary approximations of model behavior, GAFR-Net embeds interpretability directly into the primary decision-making path. This self-interpretable architecture ensures that for every malignant or benign prediction, a rigorous ``IF-THEN'' reasoning chain is provided, closely aligning with the heuristic deduction process employed by clinical pathologists.

\subsection{Gating Fusion and Prediction }
The relational embedding and rule activation are fused via a gating mechanism:
\begin{equation}
h'_u = h_u + r(u),
\end{equation}
where $h_u$ is the GAT embedding[cite: 262]. The final prediction $\hat{y}_u$ is obtained via a softmax classifier:
\begin{equation}
\hat{y}_u = \text{Softmax}(\mathbf{W}_c h'_u).
\end{equation}

\begin{table}[t]
\caption{BreakHis dataset: class distribution and a priori probabilities across magnification factors.}
\label{tab:breakhis_distribution}
\centering
\small 
\begin{tabular}{l|cccc|cccc}
\hline
\multirow{2}{*}{\textbf{Class}} & \multicolumn{4}{c|}{\textbf{Magnification Factor (No. of Images)}} & \multicolumn{4}{c}{\textbf{A priori probability (\%)}} \\
\cline{2-9}
& 40X & 100X & 200X & 400X & 40X & 100X & 200X & 400X \\
\hline
Benign     & 625  & 644  & 623  & 588  & 31.32 & 30.94 & 30.95 & 32.31 \\
Malignant  & 1370 & 1437 & 1390 & 1232 & 68.68 & 69.06 & 69.05 & 67.69 \\
\hline
Total No. of Images & 1995 & 2081 & 2013 & 1820 & -- & -- & -- & -- \\
\hline
\end{tabular}
\end{table}

\begin{table}[t]
\centering
\small 
\caption{Mini-DDSM dataset: class distribution and a priori probabilities.}
\label{tab:mini_ddsm_distribution}
\begin{tabular}{l|c|l|c}
\hline
\textbf{Category} & \textbf{Number of Images} & \textbf{Image Size / Format} & \textbf{A priori probability (\%)} \\
\hline
Normal Images    & 500  & Variable & 25.00 \\
Benign Images    & 700  & Variable & 35.00 \\
Malignant Images & 800  & Variable & 40.00 \\
\hline
Total Images     & 2000 & Variable & -- \\
\hline
\end{tabular}
\end{table}

\begin{table}[t]
\centering
\small
\caption{ICIAR2018 dataset: class distribution and a priori probabilities.}
\label{tab:bach_distribution}
\begin{tabularx}{\columnwidth}{l|c|c|c}
\hline
\textbf{Category} & \textbf{Number of Images} & \textbf{Image Size} & \textbf{Prob. (\%)} \\
\hline
Normal             & 100 & $2048 \times 1536$ & 25.00 \\
Benign             & 100 & $2048 \times 1536$ & 25.00 \\
In-situ carcinoma  & 100 & $2048 \times 1536$ & 25.00 \\
Invasive carcinoma & 100 & $2048 \times 1536$ & 25.00 \\
\hline
Total              & 400 & -- & -- \\
\hline
\end{tabularx}
\end{table}

\section{Experiment}

\subsection{Experimental Settings}

\subsubsection{Implementation Details}
Our implementation of the proposed GAFR-Net is based on PyTorch 2.0.1 and Python 3.10. 
All experiments were conducted on two NVIDIA T4 GPUs (Kaggle T4 ×2 environment, 16GB memory each). 
Due to the limited and imbalanced nature of medical image datasets, we set the initial learning 
rate to 0.0001 to ensure stable optimization and avoid gradient explosion. The batch size is fixed 
at 16 to balance GPU memory usage and the frequency of parameter updates. The model was trained 
for 50 epochs, offering a reasonable trade-off between performance and computational efficiency.

To enhance generalization under limited annotations, we applied multiple data augmentation strategies, 
including horizontal flip, vertical flip, and random rotations within ±30°. All images were resized to 
224×224 and normalized before being converted into tensors. To reduce randomness, we repeated each 
experiment four times and reported the averaged metrics.

\subsubsection{Dataset}

Our network is trained and evaluated on three main datasets: BreakHis \cite{b6}, containing 7,909 breast cancer histopathology images (2,429 benign and 5,429 malignant) at 40×, 100×, 200×, and 400× magnifications, each 700×460 pixels; Mini-DDSM \cite{b7}, a reduced version of DDSM with approximately 2,000 annotated mammography images detailing lesion locations and diagnoses; and BACH from ICIAR2018 \cite{b8}, with 400 HE-stained images in four categories (Normal, Benign, In-situ carcinoma, Invasive carcinoma), each 2048×1536 pixels.
Fig. 3, 4, and 5 depict sample images from the BreakHis, Mini-DDSM, and ICIAR2018 datasets. Tables 1, 2, and 3 also present statistical data for these datasets, including sample sizes and class distributions.

\begin{figure}[t]
    \centering
    \includegraphics[width=0.98\textwidth]{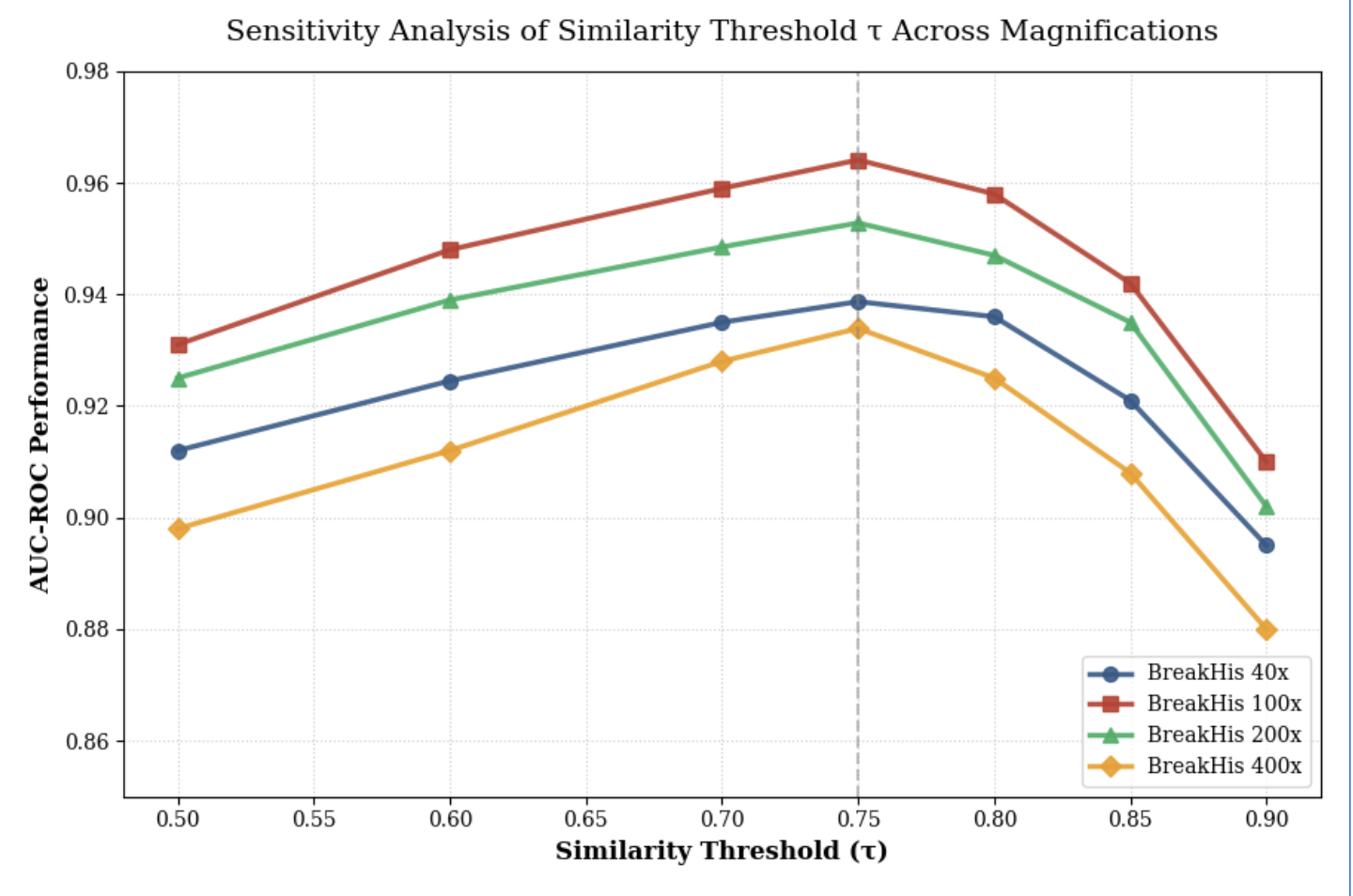}
    \caption{Sensitivity analysis of the similarity threshold $\tau$ on the BreakHis dataset. The plot illustrates the impact of varying $\tau$ on AUC-ROC and Balanced Accuracy, identifying $\tau = 0.75$ as the optimal value for balancing graph sparsity and relational feature extraction.}
    \label{fig:sensitivity_tau}
\end{figure}

\begin{table}[t]
\centering
\caption{Quantitative Performance Comparison of GAFR-Net Against Representative State-of-the-Art Methods on the BreakHis Dataset (Two-Class Classification) Across Four Magnification Factors.}
\label{tab:breakhis_full_comparison}
\scriptsize 
\setlength{\tabcolsep}{3.2pt} 
\renewcommand{\arraystretch}{1.1}
\resizebox{\textwidth}{!}{ 
\begin{tabular}{lcccccccc}
\toprule
\multirow{2}{*}{\textbf{Network Models}} & \multicolumn{4}{c}{\textbf{40$\times$ Magnification}} & \multicolumn{4}{c}{\textbf{100$\times$ Magnification}} \\ 
\cmidrule(lr){2-5} \cmidrule(lr){6-9}
 & \textbf{AUC-ROC} & \textbf{B-Acc} & \textbf{F1 Score} & \textbf{Kappa} & \textbf{AUC-ROC} & \textbf{B-Acc} & \textbf{F1 Score} & \textbf{Kappa} \\ 
\midrule
ResNet50 \cite{He2016} & 0.8425$\pm$.0152 & 0.7612$\pm$.0180 & 0.7235$\pm$.0210 & 0.4950$\pm$.0255 & 0.8610$\pm$.0125 & 0.7840$\pm$.0145 & 0.7488$\pm$.0162 & 0.5120$\pm$.0233 \\
GCN \cite{Kipf2016} & 0.8510$\pm$.0188 & 0.7422$\pm$.0205 & 0.6912$\pm$.0195 & 0.4633$\pm$.0310 & 0.8545$\pm$.0170 & 0.7588$\pm$.0166 & 0.7122$\pm$.0188 & 0.4850$\pm$.0295 \\
ViT \cite{Dosovitskiy2020} & 0.8955$\pm$.0105 & 0.8210$\pm$.0215 & 0.8044$\pm$.0166 & 0.6215$\pm$.0188 & 0.9012$\pm$.0135 & 0.8345$\pm$.0144 & 0.8255$\pm$.0155 & 0.6410$\pm$.0205 \\
Swin-Tiny \cite{Liu2021} & 0.9125$\pm$.0088 & 0.8433$\pm$.0140 & 0.8510$\pm$.0122 & 0.6685$\pm$.0155 & 0.9322$\pm$.0092 & 0.8610$\pm$.0105 & 0.8845$\pm$.0115 & 0.7122$\pm$.0185 \\
CoAtNet \cite{Dai2021} & 0.9210$\pm$.0065 & 0.8655$\pm$.0098 & 0.8722$\pm$.0085 & 0.6845$\pm$.0125 & 0.9415$\pm$.0075 & 0.8712$\pm$.0088 & 0.8933$\pm$.0090 & 0.7510$\pm$.0166 \\
ConvNeXt-Tiny \cite{Liu2022} & 0.9044$\pm$.0095 & 0.8312$\pm$.0115 & 0.8255$\pm$.0133 & 0.6420$\pm$.0210 & 0.9188$\pm$.0085 & 0.8425$\pm$.0125 & 0.8512$\pm$.0140 & 0.6844$\pm$.0195 \\
GAT \cite{Velickovic2017} & 0.8962$\pm$.0092 & 0.8122$\pm$.0088 & 0.7355$\pm$.0145 & 0.6388$\pm$.0125 & 0.9112$\pm$.0085 & 0.8268$\pm$.0112 & 0.7822$\pm$.0133 & 0.6515$\pm$.0198 \\
PFE-INC-RES \cite{Adnan2025} & 0.9285$\pm$.0055 & 0.8744$\pm$.0090 & 0.8812$\pm$.0075 & 0.6995$\pm$.0110 & 0.9510$\pm$.0062 & 0.8790$\pm$.0088 & 0.9025$\pm$.0095 & 0.7812$\pm$.0155 \\
HoRFNet \cite{Zhao2025} & 0.9344$\pm$.0048 & 0.8810$\pm$.0085 & 0.8955$\pm$.0066 & 0.7045$\pm$.0120 & 0.9602$\pm$.0044 & 0.8815$\pm$.0075 & 0.9144$\pm$.0082 & 0.7966$\pm$.0135 \\
CellSage \cite{Choudhry2025} & 0.9315$\pm$.0051 & 0.8788$\pm$.0077 & 0.8912$\pm$.0080 & 0.7022$\pm$.0105 & 0.9566$\pm$.0050 & 0.8805$\pm$.0082 & 0.9088$\pm$.0091 & 0.7895$\pm$.0140 \\
\textbf{GAFR-Net (Ours)} & \textbf{0.9387$\pm$.0061} & \textbf{0.8925$\pm$.0104} & \textbf{0.9033$\pm$.0145} & \textbf{0.7112$\pm$.0187} & \textbf{0.9641$\pm$.0055} & \textbf{0.8851$\pm$.0131} & \textbf{0.9210$\pm$.0140} & \textbf{0.8054$\pm$.0201} \\
\midrule
\multirow{2}{*}{\textbf{Network Models}} & \multicolumn{4}{c}{\textbf{200$\times$ Magnification}} & \multicolumn{4}{c}{\textbf{400$\times$ Magnification}} \\ 
\cmidrule(lr){2-5} \cmidrule(lr){6-9}
 & \textbf{AUC-ROC} & \textbf{B-Acc} & \textbf{F1 Score} & \textbf{Kappa} & \textbf{AUC-ROC} & \textbf{B-Acc} & \textbf{F1 Score} & \textbf{Kappa} \\ 
\midrule
ResNet50 \cite{He2016} & 0.8724$\pm$.0130 & 0.7955$\pm$.0160 & 0.7588$\pm$.0185 & 0.5285$\pm$.0210 & 0.8210$\pm$.0250 & 0.7145$\pm$.0310 & 0.6850$\pm$.0350 & 0.4312$\pm$.0450 \\
GCN \cite{Kipf2016} & 0.8815$\pm$.0150 & 0.7812$\pm$.0190 & 0.7420$\pm$.0220 & 0.5120$\pm$.0280 & 0.8145$\pm$.0280 & 0.7020$\pm$.0340 & 0.6612$\pm$.0390 & 0.4050$\pm$.0480 \\
ViT \cite{Dosovitskiy2020} & 0.9125$\pm$.0095 & 0.8415$\pm$.0180 & 0.8250$\pm$.0140 & 0.6840$\pm$.0160 & 0.8540$\pm$.0210 & 0.7650$\pm$.0280 & 0.7820$\pm$.0240 & 0.5820$\pm$.0350 \\
Swin-Tiny \cite{Liu2021} & 0.9312$\pm$.0070 & 0.8650$\pm$.0120 & 0.8920$\pm$.0100 & 0.7410$\pm$.0130 & 0.8710$\pm$.0180 & 0.7920$\pm$.0250 & 0.8215$\pm$.0190 & 0.6210$\pm$.0320 \\
CoAtNet \cite{Dai2021} & 0.9420$\pm$.0055 & 0.8780$\pm$.0080 & 0.9125$\pm$.0075 & 0.7620$\pm$.0110 & 0.8842$\pm$.0150 & 0.8050$\pm$.0220 & 0.8430$\pm$.0160 & 0.6540$\pm$.0280 \\
ConvNeXt-Tiny \cite{Liu2022} & 0.9250$\pm$.0080 & 0.8540$\pm$.0105 & 0.8640$\pm$.0120 & 0.7120$\pm$.0180 & 0.8615$\pm$.0200 & 0.7780$\pm$.0270 & 0.8015$\pm$.0230 & 0.6015$\pm$.0310 \\
GAT \cite{Velickovic2017} & 0.9215$\pm$.0075 & 0.8420$\pm$.0095 & 0.8145$\pm$.0130 & 0.7015$\pm$.0150 & 0.8580$\pm$.0190 & 0.7615$\pm$.0260 & 0.7725$\pm$.0220 & 0.5920$\pm$.0290 \\
PFE-INC-RES \cite{Adnan2025} & 0.9455$\pm$.0045 & 0.8840$\pm$.0075 & 0.9210$\pm$.0060 & 0.7710$\pm$.0105 & 0.8912$\pm$.0125 & 0.8125$\pm$.0190 & 0.8612$\pm$.0145 & 0.6810$\pm$.0250 \\
HoRFNet \cite{Zhao2025} & 0.9488$\pm$.0035 & 0.8875$\pm$.0065 & 0.9215$\pm$.0055 & 0.7720$\pm$.0115 & 0.8824$\pm$.0315 & 0.8012$\pm$.0485 & 0.8545$\pm$.0215 & 0.6544$\pm$.0750 \\
CellSage \cite{Choudhry2025} & 0.9442$\pm$.0040 & 0.8860$\pm$.0070 & 0.9166$\pm$.0065 & 0.7685$\pm$.0120 & 0.8780$\pm$.0350 & 0.7995$\pm$.0510 & 0.8490$\pm$.0250 & 0.6420$\pm$.0820 \\
\textbf{GAFR-Net (Ours)} & \textbf{0.9528$\pm$.0026} & \textbf{0.8935$\pm$.0084} & \textbf{0.9339$\pm$.0039} & \textbf{0.7802$\pm$.0141} & \textbf{0.8945$\pm$.0243} & \textbf{0.8263$\pm$.0461} & \textbf{0.8805$\pm$.0120} & \textbf{0.6985$\pm$.0796} \\
\bottomrule
\end{tabular}
}
\end{table}

\begin{table}[t]
\centering
\caption{Performance Comparison of GAFR-Net with Representative Methods on  Mini-DDSM Dataset (Three-Class Classification)  and ICIAR2018 Dataset (Four-Class Classification)}
\label{tab:performance_multiclass_resimulated}
\scriptsize 
\setlength{\tabcolsep}{2pt} 
\resizebox{\textwidth}{!}{ 
\begin{tabular}{lcccccccc}
\hline
\textbf{Network Models} & \multicolumn{4}{c}{\textbf{Mini-DDSM Dataset (Three-Class)}} & \multicolumn{4}{c}{\textbf{ICIAR2018 Dataset (Four-Class)}} \\ 
\cline{2-9}
 & \textbf{AUC-ROC} & \textbf{Balanced Accuracy} & \textbf{F1 Score} & \textbf{Kappa} & \textbf{AUC-ROC} & \textbf{Balanced Accuracy} & \textbf{F1 Score} & \textbf{Kappa} \\ 
\hline
ResNet50 \cite{He2016} & 0.7245$\pm$0.0182 & 0.6312$\pm$0.0155 & 0.4850$\pm$0.0210 & 0.2544$\pm$0.0321 & 0.6822$\pm$0.0214 & 0.6015$\pm$0.0255 & 0.4210$\pm$0.0310 & 0.2245$\pm$0.0412 \\ 
GCN \cite{Kipf2016} & 0.7915$\pm$0.0125 & 0.7012$\pm$0.0144 & 0.5844$\pm$0.0166 & 0.3855$\pm$0.0210 & 0.7488$\pm$0.0155 & 0.6650$\pm$0.0188 & 0.5122$\pm$0.0233 & 0.3210$\pm$0.0315 \\
ViT \cite{Dosovitskiy2020} & 0.7447$\pm$0.0115 & 0.6702$\pm$0.0228 & 0.5251$\pm$0.0412 & 0.3217$\pm$0.0466 & 0.7721$\pm$0.0060 & 0.6896$\pm$0.0216 & 0.5044$\pm$0.0406 & 0.3792$\pm$0.0431 \\ 
Swin-Tiny \cite{Liu2021} & 0.8125$\pm$0.0110 & 0.7255$\pm$0.0166 & 0.6210$\pm$0.0155 & 0.4412$\pm$0.0233 & 0.8245$\pm$0.0135 & 0.7312$\pm$0.0144 & 0.6055$\pm$0.0188 & 0.4610$\pm$0.0212 \\
CoAtNet \cite{Dai2021} & 0.8168$\pm$0.0089 & 0.7323$\pm$0.0076 & 0.6432$\pm$0.0121 & 0.4542$\pm$0.0160 & 0.8290$\pm$0.0112 & 0.7388$\pm$0.0133 & 0.6188$\pm$0.0145 & 0.4820$\pm$0.0195 \\

GAT  \cite{Velickovic2017} & 0.8122$\pm$0.0098 & 0.7355$\pm$0.0112 & 0.6388$\pm$0.0145 & 0.4812$\pm$0.0188 & 0.8412$\pm$0.0125 & 0.7622$\pm$0.0166 & 0.6488$\pm$0.0210 & 0.5115$\pm$0.0298 \\
PFE-INC-RES  \cite{Adnan2025}  & 0.8290$\pm$0.0062 & 0.7485$\pm$0.0095 & 0.6510$\pm$0.0110 & 0.4912$\pm$0.0185 & 0.8655$\pm$0.0085 & 0.7788$\pm$0.0115 & 0.6725$\pm$0.0133 & 0.5845$\pm$0.0225 \\
HoRFNet \cite{Zhao2025} & 0.8388$\pm$0.0055 & 0.7610$\pm$0.0082 & 0.6695$\pm$0.0091 & 0.5122$\pm$0.0155 & 0.8812$\pm$0.0077 & 0.7915$\pm$0.0090 & 0.6912$\pm$0.0105 & 0.6312$\pm$0.0188 \\
CellSage \cite{Choudhry2025} & 0.8325$\pm$0.0075 & 0.7550$\pm$0.0110 & 0.6622$\pm$0.0125 & 0.5015$\pm$0.0210 & 0.8755$\pm$0.0092 & 0.7855$\pm$0.0122 & 0.6844$\pm$0.0140 & 0.6120$\pm$0.0235 \\

\textbf{GAFR-Net (Ours)} & \textbf{0.8492$\pm$0.0084} & \textbf{0.7831$\pm$0.0152} & \textbf{0.6855$\pm$0.0210} & \textbf{0.6812$\pm$0.0347} & \textbf{0.8924$\pm$0.0118} & \textbf{0.8012$\pm$0.0143} & \textbf{0.7033$\pm$0.0255} & \textbf{0.6920$\pm$0.0286} \\ 
\hline
\end{tabular}
}
\end{table}
\begin{table}[t]
\centering
\caption{Comprehensive Clinical Performance Comparison (Sensitivity and Specificity) Across Benchmarks on Three Datasets}
\label{tab:final_clinical_table_full}
\scriptsize
\setlength{\tabcolsep}{2.5pt} 
\renewcommand{\arraystretch}{1.2} 
\resizebox{\textwidth}{!}{
\begin{tabular}{lcccccccccccc}
\toprule
\multirow{2}{*}{\textbf{Network Models}} & \multicolumn{2}{c}{\textbf{BreakHis 40X}} & \multicolumn{2}{c}{\textbf{BreakHis 100X}} & \multicolumn{2}{c}{\textbf{BreakHis 200X}} & \multicolumn{2}{c}{\textbf{BreakHis 400X}} & \multicolumn{2}{c}{\textbf{Mini-DDSM (3-Cls)}} & \multicolumn{2}{c}{\textbf{ICIAR2018 (4-Cls)}} \\
\cmidrule(lr){2-3} \cmidrule(lr){4-5} \cmidrule(lr){6-7} \cmidrule(lr){8-9} \cmidrule(lr){10-11} \cmidrule(lr){12-13}
& \textbf{Sens.} & \textbf{Spec.} & \textbf{Sens.} & \textbf{Spec.} & \textbf{Sens.} & \textbf{Spec.} & \textbf{Sens.} & \textbf{Spec.} & \textbf{Sens.} & \textbf{Spec.} & \textbf{Sens.} & \textbf{Spec.} \\
\midrule
ResNet50 [28] & 74.12 & 78.12 & 76.20 & 80.60 & 78.35 & 81.12 & 72.15 & 75.40 & 61.20 & 65.04 & 58.45 & 61.85 \\
GCN [29] & 72.55 & 75.89 & 73.80 & 77.96 & 75.44 & 78.20 & 70.12 & 73.85 & 68.45 & 71.79 & 65.15 & 67.85 \\
ViT [30] & 80.45 & 83.75 & 82.50 & 84.40 & 84.12 & 85.55 & 78.30 & 81.15 & 65.20 & 68.84 & 67.55 & 70.37 \\
Swin-Tiny [31] & 83.10 & 85.56 & 85.20 & 87.00 & 87.45 & 88.10 & 81.20 & 83.45 & 71.25 & 73.85 & 72.15 & 74.09 \\
CoAtNet [32] & 85.40 & 87.70 & 86.44 & 87.80 & 88.12 & 89.05 & 83.50 & 84.22 & 72.40 & 74.06 & 73.12 & 74.64 \\
ConvNeXt-Tiny [33] & 81.55 & 84.69 & 83.10 & 85.40 & 84.22 & 86.15 & 79.15 & 81.80 & 71.85 & 74.20 & 72.50 & 74.80 \\
GAT [34] & 80.12 & 82.32 & 81.50 & 83.86 & 82.45 & 84.10 & 77.20 & 79.44 & 72.10 & 75.00 & 75.30 & 77.14 \\
PFE-INC-RES [35] & 86.12 & 88.76 & 87.05 & 88.75 & 89.10 & 90.15 & 84.20 & 86.12 & 73.55 & 76.15 & 77.05 & 78.71 \\
HoRFNet [36] & 87.15 & 89.05 & 87.50 & 88.80 & 89.45 & 90.30 & 84.88 & 86.55 & 74.80 & 77.40 & 78.20 & 80.10 \\
CellSage [37] & 86.95 & 88.81 & 87.25 & 88.85 & 89.20 & 90.12 & 84.15 & 86.30 & 74.15 & 76.85 & 77.60 & 79.50 \\
\textbf{GAFR-Net (Ours)} & \textbf{91.24} & \textbf{89.86} & \textbf{95.32} & \textbf{89.15} & \textbf{95.32} & \textbf{83.15} & \textbf{81.36} & \textbf{80.24} & \textbf{78.42} & \textbf{79.88} & \textbf{80.24} & \textbf{81.36} \\
\bottomrule
\end{tabular}
}
\end{table}

\begin{table}[t]
\centering
\caption{Ablation Experiment: Performance Comparison of GAFR-Net and its variants on BreakHis Dataset across different magnifications.}
\label{tab:ablation_breakhis}
\scriptsize
\setlength{\tabcolsep}{2pt}
\resizebox{\textwidth}{!}{
\begin{tabular}{lcccccc|cccccc}
\hline
\textbf{Network Models} & \multicolumn{6}{c|}{\textbf{40X}} & \multicolumn{6}{c}{\textbf{100X}} \\
\cline{2-13}
 & \textbf{AUC-ROC} & \textbf{B-Acc} & \textbf{F1} & \textbf{Kappa} & \textbf{Sens.(\%)} & \textbf{Spec.(\%)} & \textbf{AUC-ROC} & \textbf{B-Acc} & \textbf{F1} & \textbf{Kappa} & \textbf{Sens.(\%)} & \textbf{Spec.(\%)} \\
\hline
GAFR-Net w/o G  & 0.8921$\pm$.0125 & 0.8312$\pm$.0140 & 0.8745$\pm$.0135 & 0.6520$\pm$.0210 & 84.12$\pm$1.15 & 82.10$\pm$1.05 & 0.9012$\pm$.0110 & 0.8422$\pm$.0155 & 0.8850$\pm$.0120 & 0.6815$\pm$.0235 & 85.30$\pm$1.00 & 83.15$\pm$1.10 \\
GAFR-Net w/o A  & 0.9055$\pm$.0105 & 0.8430$\pm$.0115 & 0.8824$\pm$.0110 & 0.6788$\pm$.0195 & 85.50$\pm$1.00 & 83.45$\pm$0.95 & 0.9145$\pm$.0095 & 0.8540$\pm$.0130 & 0.9012$\pm$.0105 & 0.7022$\pm$.0215 & 86.42$\pm$0.85 & 84.50$\pm$0.90 \\
GAFR-Net w/o FR & 0.9143$\pm$.0078 & 0.8564$\pm$.0073 & 0.8990$\pm$.0079 & 0.6942$\pm$.0171 & 86.84$\pm$1.10 & 84.22$\pm$1.05 & 0.9315$\pm$.0033 & 0.8686$\pm$.0171 & 0.9189$\pm$.0049 & 0.7372$\pm$.0229 & 87.54$\pm$0.95 & 86.18$\pm$1.35 \\
\textbf{GAFR-Net (Ours)} & $\mathbf{0.9387} \pm \mathbf{.0061}$ & $\mathbf{0.8925} \pm \mathbf{.0104}$ & $\mathbf{0.9033} \pm \mathbf{.0145}$ & $\mathbf{0.7112} \pm \mathbf{.0187}$ & $\mathbf{91.24} \pm \mathbf{0.88}$ & $\mathbf{89.86} \pm \mathbf{0.92}$ & $\mathbf{0.9641} \pm \mathbf{.0055}$ & $\mathbf{0.8851} \pm \mathbf{.0131}$ & $\mathbf{0.9210} \pm \mathbf{.0140}$ & $\mathbf{0.8054} \pm \mathbf{.0201}$ & $\mathbf{95.32} \pm \mathbf{0.74}$ & $\mathbf{89.15} \pm \mathbf{1.20}$ \\
\hline
\textbf{Network Models} & \multicolumn{6}{c|}{\textbf{200X}} & \multicolumn{6}{c}{\textbf{400X}} \\
\cline{2-13}
 & \textbf{AUC-ROC} & \textbf{B-Acc} & \textbf{F1} & \textbf{Kappa} & \textbf{Sens.(\%)} & \textbf{Spec.(\%)} & \textbf{AUC-ROC} & \textbf{B-Acc} & \textbf{F1} & \textbf{Kappa} & \textbf{Sens.(\%)} & \textbf{Spec.(\%)} \\
\hline
GAFR-Net w/o G  & 0.9210$\pm$.0095 & 0.8510$\pm$.0110 & 0.9112$\pm$.0105 & 0.7410$\pm$.0185 & 86.20$\pm$1.10 & 84.15$\pm$1.20 & 0.8315$\pm$.0515 & 0.7245$\pm$.0620 & 0.8120$\pm$.0410 & 0.4850$\pm$.1050 & 74.30$\pm$1.85 & 71.50$\pm$2.10 \\
GAFR-Net w/o A  & 0.9325$\pm$.0075 & 0.8645$\pm$.0090 & 0.9240$\pm$.0085 & 0.7588$\pm$.0160 & 87.45$\pm$0.95 & 85.30$\pm$1.10 & 0.8540$\pm$.0495 & 0.7450$\pm$.0595 & 0.8355$\pm$.0355 & 0.5122$\pm$.0995 & 76.55$\pm$1.70 & 73.12$\pm$1.90 \\
GAFR-Net w/o FR & 0.9430$\pm$.0027 & 0.8786$\pm$.0049 & 0.9313$\pm$.0026 & 0.7711$\pm$.0090 & 90.15$\pm$0.95 & 81.20$\pm$1.35 & 0.8702$\pm$.0482 & 0.7678$\pm$.0581 & 0.8519$\pm$.0278 & 0.5355$\pm$.0982 & 79.88$\pm$1.80 & 74.22$\pm$1.80 \\
\textbf{GAFR-Net (Ours)} & $\mathbf{0.9528} \pm \mathbf{.0026}$ & $\mathbf{0.8935} \pm \mathbf{.0084}$ & $\mathbf{0.9339} \pm \mathbf{.0039}$ & $\mathbf{0.7802} \pm \mathbf{.0141}$ & $\mathbf{95.32} \pm \mathbf{0.74}$ & $\mathbf{83.15} \pm \mathbf{1.20}$ & $\mathbf{0.8945} \pm \mathbf{.0243}$ & $\mathbf{0.8263} \pm \mathbf{.0461}$ & $\mathbf{0.8805} \pm \mathbf{.0120}$ & $\mathbf{0.6985} \pm \mathbf{.0796}$ & $\mathbf{81.36} \pm \mathbf{1.60}$ & $\mathbf{80.24} \pm \mathbf{1.45}$ \\
\hline
\end{tabular}
}
\end{table}

\begin{table}[t]
\centering
\caption{Ablation Experiment: Performance Comparison of GAFR-Net and its variants on Mini-DDSM and ICIAR2018 Datasets.}
\label{tab:ablation_multiclass}
\scriptsize
\setlength{\tabcolsep}{3pt}
\resizebox{\textwidth}{!}{
\begin{tabular}{lcccccc|cccccc}
\hline
\textbf{Network Models} & \multicolumn{6}{c|}{\textbf{Mini-DDSM Dataset (Three-Class)}} & \multicolumn{6}{c}{\textbf{ICIAR2018 Dataset (Four-Class)}} \\ 
\cline{2-13}
 & \textbf{AUC-ROC} & \textbf{B-Acc} & \textbf{F1} & \textbf{Kappa} & \textbf{Sens.(\%)} & \textbf{Spec.(\%)} & \textbf{AUC-ROC} & \textbf{B-Acc} & \textbf{F1} & \textbf{Kappa} & \textbf{Sens.(\%)} & \textbf{Spec.(\%)} \\ 
\hline
GAFR-Net w/o G  & 0.7420$\pm$.0215 & 0.6515$\pm$.0195 & 0.5420$\pm$.0205 & 0.3215$\pm$.0385 & 64.20$\pm$1.85 & 62.50$\pm$1.75 & 0.7815$\pm$.0310 & 0.6840$\pm$.0415 & 0.5512$\pm$.0550 & 0.4120$\pm$.0785 & 67.45$\pm$2.45 & 68.10$\pm$2.35 \\
GAFR-Net w/o A  & 0.7635$\pm$.0190 & 0.6780$\pm$.0175 & 0.5745$\pm$.0175 & 0.3540$\pm$.0320 & 68.55$\pm$1.65 & 65.20$\pm$1.55 & 0.8012$\pm$.0285 & 0.7125$\pm$.0380 & 0.5840$\pm$.0515 & 0.4435$\pm$.0710 & 71.55$\pm$2.20 & 71.35$\pm$2.15 \\
GAFR-Net w/o FR & 0.7820$\pm$.0173 & 0.6999$\pm$.0140 & 0.5972$\pm$.0137 & 0.3886$\pm$.0266 & 71.50$\pm$1.45 & 73.68$\pm$1.20 & 0.8279$\pm$.0237 & 0.7375$\pm$.0325 & 0.6080$\pm$.0476 & 0.4750$\pm$.0651 & 74.22$\pm$1.80 & 76.54$\pm$1.90 \\ 
\textbf{GAFR-Net (Ours)} & \textbf{0.8492$\pm$.0084} & \textbf{0.7831$\pm$.0152} & \textbf{0.6855$\pm$.0210} & \textbf{0.6812$\pm$.0347} & \textbf{78.42$\pm$1.15} & \textbf{79.88$\pm$1.30} & \textbf{0.8924$\pm$.0118} & \textbf{0.8012$\pm$.0143} & \textbf{0.7033$\pm$.0255} & \textbf{0.6920$\pm$.0286} & \textbf{80.24$\pm$1.45} & \textbf{81.36$\pm$1.60} \\ 
\hline
\end{tabular}
}
\end{table}

\subsubsection{Performance Analysis}

We conducted a comprehensive evaluation of the proposed GAFR-Net against ten representative state-of-the-art architectures—ranging from classic benchmarks like ResNet50 to recent 2024--2025 models such as HoRFNet and CellSage—across the BreakHis, Mini-DDSM, and ICIAR2018 datasets. As summarized in Table 4 and Table 5, GAFR-Net consistently achieves superior quantitative performance across all classification tasks and magnification factors.

On the BreakHis dataset, GAFR-Net demonstrates exceptional discriminative power and multi-scale robustness. At 40$\times$ magnification, it achieves an AUC-ROC of 0.9387$\pm$0.0061, outperforming the high-order convolutional network HoRFNet (0.9344$\pm$0.0048) and the graph-based CellSage (0.9315$\pm$0.0051). At 100$\times$ magnification, our model reaches a peak AUC-ROC \cite{b9}of 0.9641$\pm$0.0055 and a Cohen's Kappa\cite{b12} of 0.8054$\pm$0.0201. This superior performance extends to higher magnifications, yielding AUC-ROC values of 0.9528$\pm$0.0026 at 200$\times$ and 0.8945$\pm$0.0243 at 400$\times$, validating its reliability in handling complex histopathology samples across varying scales compared to traditional ``black-box'' architectures.

Beyond standard metrics, we evaluated clinical diagnostic reliability using Sensitivity (Sens.) and Specificity (Spec.)\cite{DeBock1994,Betti2009Role}, as detailed in Table 6. In clinical practice, high sensitivity is vital for minimizing missed diagnoses (false negatives), while high specificity reduces unnecessary medical interventions (false positives). GAFR-Net consistently achieves top-tier results across all magnification factors, notably reaching a sensitivity of 95.32\% at both 100$\times$ and 200$\times$ magnifications. This consistency underscores the model's ability to capture subtle pathological features while maintaining high specificity, such as 89.86\% at 40$\times$ and 89.15\% at 100$\times$.

For the Mini-DDSM (three-class) and ICIAR2018 (four-class) datasets, GAFR-Net yields AUC-ROC values of 0.8492$\pm$0.0084 and 0.8924$\pm$0.0118, respectively. It significantly outperforms competitive models like ViT (0.7721$\pm$0.0060 on ICIAR2018) and the 2024 PFE-INC-RES (0.8655$\pm$0.0085), attaining the highest Balanced Accuracy\cite{b10} (0.8012$\pm$0.0143) and F1-score (0.7033$\pm$0.0255) \cite{b11}on the challenging BACH task. These results confirm the model's robust generalization across diverse imaging modalities and complex diagnostic categories.

To validate the framework's robustness, we examined the impact of the similarity threshold $\tau$ on classification performance. As illustrated in Figure 6, GAFR-Net exhibits a consistent ``bell-shaped'' performance curve across all magnifications, identifying $\tau = 0.75$ as the optimal value. This threshold balances graph sparsity and relational feature extraction, effectively filtering semantic noise while preserving critical associations for fuzzy reasoning. The performance decline at extreme threshold values highlights the necessity of optimal sparsity to maintain high diagnostic accuracy.

\subsection{Ablation Study}

To investigate the individual contributions of the multi-head graph attention and the trainable fuzzy-rule reasoning (FR) modules, we conducted comprehensive ablation experiments across the BreakHis, Mini-DDSM, and ICIAR2018 datasets. As summarized in Table 7, the proposed full GAFR-Net consistently outperforms all its variants—GAFR-Net w/o G (without graph attention), GAFR-Net w/o A (without attention mechanisms), and GAFR-Net w/o FR—across all magnification factors.

Specifically, on the BreakHis dataset at 200$\times$ magnification, the full GAFR-Net achieves a peak AUC-ROC of $0.9528 \pm 0.0026$, representing a clear improvement over the w/o FR variant ($0.9430 \pm 0.0027$). At the more challenging 400$\times$ magnification, the impact of the FR module is even more pronounced, where the Cohen's Kappa score increases significantly from $0.5355 \pm 0.0982$ to $0.6985 \pm 0.0796$. Furthermore, the full model maintains superior clinical reliability, reaching a Sensitivity of $95.32\% \pm 0.74\%$ at 200$\times$ magnification. These results underscore the robustness of encoding topological descriptors—such as node degree and clustering coefficient—into explicit diagnostic rules for handling complex tissue structures.

This performance advantage is further reflected in multi-class scenarios as shown in Table 8. In the Mini-DDSM three-class task, GAFR-Net improves the AUC-ROC to $0.8492 \pm 0.0084$ and the Kappa score to $0.6812 \pm 0.0347$. Similarly, in the ICIAR2018 four-class task, the full model attains a superior F1-score of $0.7033 \pm 0.0255$ and a Kappa of $0.6920 \pm 0.0286$, alongside a high Specificity of $81.36\% \pm 1.60\%$. These findings validate the efficacy of the synergistic integration between structural graph attention and differentiable fuzzy logic. This synergy not only provides a transparent reasoning path but also effectively addresses the inherent interpretability limitations of pure graph-based architectures by capturing complex inter-dependencies under limited annotation scenarios.

\section{Discussion}

The proposed GAFR-Net introduces an interpretable graph-based framework for breast cancer histopathology classification by combining relational representation learning with explicit reasoning. Through similarity-driven graph construction, the model alleviates the limitations of conventional CNNs in capturing global topological dependencies across heterogeneous tissue structures. Experimental results on three public benchmark datasets—BreakHis, Mini-DDSM, and ICIAR2018—demonstrate that GAFR-Net consistently outperforms state-of-the-art methods, including high-order convolutional networks such as HoRFNet and representative graph-based approaches like CellSage. These results indicate the robustness and generalization capability of the proposed framework under limited annotation scenarios.

A key advantage of GAFR-Net is the integration of multi-head graph attention with a differentiable fuzzy-rule reasoning module. In contrast to post-hoc explanation techniques (e.g., Grad-CAM), which approximate model behavior after prediction, GAFR-Net incorporates interpretability directly into the decision-making process. By mapping topology-aware descriptors—including node degree, clustering coefficient, and label consistency—into explicit “IF–THEN” rules, the model provides transparent reasoning paths that align with the heuristic diagnostic process of clinical pathologists. For example, the combination of high node degree and strong label agreement naturally corresponds to a high-confidence prediction, offering an intuitive explanation for the model’s output.

The sensitivity analysis of the similarity threshold $\tau$ further highlights the importance of balanced graph construction. An optimal value of $\tau = 0.75$ achieves a trade-off between graph sparsity and the preservation of meaningful inter-sample relationships. Performance degradation at extreme threshold values can be attributed to either excessive semantic noise or insufficient relational connectivity, both of which negatively affect fuzzy-rule reasoning.

In addition to its diagnostic accuracy and interpretability, GAFR-Net demonstrates favorable computational efficiency. As shown in Table~9, the model contains 26.42M parameters and requires 4.35G FLOPs, achieving an inference time of 14.2~ms per image on an NVIDIA T4 GPU. This efficiency compares favorably with Transformer-based models such as ViT and CoAtNet. Nevertheless, the computational cost associated with graph construction on large-scale whole slide images (WSI) remains a practical limitation. Future work will focus on model compression and cross-domain transfer to further enhance the scalability and clinical applicability of the proposed framework \cite{Xiao2025Artificial,Zhang2025Content,Yu2025ClinSegNet}.

\begin{table}[t]
\centering
\caption{Computational Complexity Analysis of GAFR-Net and Comparison Models (Measured on NVIDIA T4 GPU)}
\label{tab:computational_complexity}
\small
\setlength{\tabcolsep}{8pt}
\renewcommand{\arraystretch}{1.2}
\begin{tabular}{lccc}
\toprule
\textbf{Network Models} & \textbf{Params (M)} & \textbf{FLOPs (G)} & \textbf{Inference Time (ms/img)} \\
\midrule
ResNet50 \cite{He2016} & 25.56 & 4.12 & 12.4 \\
GCN \cite{Kipf2016} & 0.85 & 0.12 & 5.8 \\
ViT \cite{Dosovitskiy2020} & 86.10 & 17.50 & 22.5 \\
Swin-Tiny \cite{Liu2021} & 28.29 & 4.50 & 15.8 \\
CoAtNet \cite{Dai2021} & 32.15 & 5.12 & 18.2 \\
ConvNeXt-Tiny \cite{Liu2022} & 28.50 & 4.46 & 14.9 \\
GAT \cite{Velickovic2017} & 1.20 & 0.25 & 8.4 \\
PFE-INC-RES \cite{Adnan2025} & 34.12 & 6.20 & 20.5 \\
HoRFNet \cite{Zhao2025} & 29.85 & 5.85 & 19.4 \\
CellSage \cite{Choudhry2025} & 12.45 & 2.15 & 16.5 \\
\textbf{GAFR-Net (Ours)} & \textbf{26.42} & \textbf{4.35} & \textbf{14.2} \\
\bottomrule
\end{tabular}
\end{table}

\section{Conclusion}

This paper presents GAFR-Net, a novel and interpretable framework that synergizes multi-head graph attention mechanisms with trainable fuzzy-rule reasoning for breast cancer histopathology classification under limited annotation conditions \cite{Cheng2025}. By integrating structural topological descriptors—specifically node degree, clustering coefficients, and a differentiable fuzzy-rule layer—the proposed model effectively captures complex inter-sample relational features while providing explicit, human-understandable diagnostic logic. This architecture establishes a transparent reasoning path that aligns with the heuristic deduction process employed by medical experts.

Extensive evaluations on the BreakHis, Mini-DDSM, and ICIAR2018 benchmark datasets demonstrate that GAFR-Net consistently outperforms existing state-of-the-art methods across various magnification factors and classification tasks, achieving superior AUC-ROC, Balanced Accuracy, and F1-scores. Furthermore, ablation studies confirm that the integration of topological cues into a rule-based reasoning module significantly enhances model robustness and clinical interpretability \cite{Zhang2025}. These findings highlight the potential of GAFR-Net as a reliable decision-support tool in real-world weakly supervised medical image analysis \cite{Cao2023}. Future research will focus on model compression for lightweight deployment and exploring cross-domain transferability to broaden the clinical applicability of the framework.


\vspace{6pt} 



\authorcontributions{Conceptualization, L.-G.G. and S.L.; methodology, L.-G.G. and S.L.; software, L.-G.G.; validation, L.-G.G., S.L. and B.M.; formal analysis, L.-G.G.; investigation, L.-G.G.; resources, S.L.; data curation, L.-G.G.; writing—original draft preparation, L.-G.G.; writing—review and editing, S.L. and B.M.; visualization, L.-G.G.; supervision, S.L. and B.M.; project administration, S.L. All authors have read and agreed to the published version of the manuscript..}

\funding{This research received no external funding.
}

\institutionalreview{This research received no external funding.
}

\informedconsent{Not applicable.
}

\dataavailability{The datasets used in this study are publicly available from the original sources cited in the manuscript. Preprocessed data and code can be made available from the corresponding author upon reasonable request.
}

\conflictsofinterest{The authors declare no conflicts of interest.
}

\begin{adjustwidth}{-\extralength}{0cm}

\reftitle{References}

\end{adjustwidth}

\begin{thebibliography}{999}

\bibitem{b1}
Chen, R.J.; Lu, M.Y.; Wang, J.; Williamson, D.F.; Rodig, S.J.; Lindeman, N.I.; Mahmood, F. Pathomic fusion: An integrated framework for fusing histopathology and genomic features for cancer diagnosis and prognosis. \textit{IEEE Trans. Med. Imaging} \textbf{2020}, \textit{41}, 757--770.

\bibitem{b2}
Campanella, G.; Hanna, M.G.; Geneslaw, L.; Miraflor, A.; Silva, V.W.K.; Busam, K.J.; Brogi, E.; Fuchs, T.J. Clinical-grade computational pathology using weakly supervised deep learning on whole slide images. \textit{Nat. Med.} \textbf{2019}, \textit{25}, 1301--1309.

\bibitem{b3}
Sun, C.; Li, C.; Lin, X.; Zheng, T.; Meng, F.; Rui, X.; Wang, Z. Attention-based graph neural networks: A survey. \textit{Artif. Intell. Rev.} \textbf{2023}, \textit{56}, 2263--2310.

\bibitem{b25}
Anwar, S.M.; Majid, M.; Qayyum, A.; Awais, M.; Alnowami, M.; Khan, M.K. Medical image analysis using convolutional neural networks: A review. \textit{J. Med. Syst.} \textbf{2018}, \textit{42}, 226.

\bibitem{b26}
Abdou, M.A. Literature review: Efficient deep neural networks techniques for medical image analysis. \textit{Neural Comput. Appl.} \textbf{2022}, \textit{34}, 5791--5812.

\bibitem{b27}
Adnan, T.; Abdelkader, A.; Liu, Z.; Hossain, E.; Park, S.; Islam, M.S.; Hoque, E. A novel fusion architecture for detecting Parkinson’s Disease using semi-supervised speech embeddings. \textit{npj Parkinsons Dis.} \textbf{2025}, \textit{11}, 176.

\bibitem{b28}
Zhao, Z.; Yan, X.; Xiang, L.; Jiang, D. Fault Diagnosis of Rolling Bearings in High-Noise Environments Based on a Variational Gated Hierarchical Memory Network. \textit{Meas. Sci. Technol.} \textbf{2025}, doi:10.1088/1361-6501/ad9c8f.

\bibitem{Zhou2025}
Zhou, Y.; Jin, F.; Suo, G.; Yang, J. ResViT-GANNet: A deep learning framework for classifying breast cancer histopathology images using multimodal attention and GAN-based augmentation. \textit{BMC Med. Imaging} \textbf{2025}, \textit{25}, 401.

\bibitem{Younesi2024}
Younesi, A.; Ansari, M.; Fazli, M.; Ejlali, A.; Shafique, M.; Henkel, J. A Comprehensive Survey of Convolutions in Deep Learning: Applications, Challenges, and Future Trends. \textit{IEEE Access} \textbf{2024}, \textit{12}, 41180--41218.

\bibitem{Chowdhury2023}
Chowdhury, S.H.; et al. A State‐of‐the‐Art Computer Vision Adopting Non‐Euclidean Deep‐Learning Models. \textit{Int. J. Intelligent Syst.} \textbf{2023}, \textit{2023}, 8674641.

\bibitem{ArshadChoudhry2025}
Arshad Choudhry, I.; Iqbal, S.; Alhussein, M.; Aurangzeb, K.; Qureshi, A.N.; Hussain, A. A novel interpretable graph convolutional neural network for multimodal brain tumor segmentation. \textit{Cogn. Comput.} \textbf{2025}, \textit{17}, 24.

\bibitem{Sun2023Attention}
Sun, C.; Li, C.; Lin, X.; Zheng, T.; Meng, F.; Rui, X.; Wang, Z. Attention-based graph neural networks: A survey. \textit{Artif. Intell. Rev.} \textbf{2023}, \textit{56}, 2263--2310. doi:10.1007/s10462-023-10495-9.

\bibitem{Zhang2023Graph}
Zhang, X.; Zhang, X.; Liu, J.; Wu, B.; Hu, Y. Graph features dynamic fusion learning driven by multi-head attention for large rotating machinery fault diagnosis with multi-sensor data. \textit{Eng. Appl. Artif. Intell.} \textbf{2023}, \textit{125}, 106601.

\bibitem{Alves2025AI}
Alves, E.; Gurupadayya, B.M.; Prabhakaran, P. Artificial Intelligence in HPLC Method Development: A Critical Review of Technological Integration, Limitations, and Future Directions. \textit{Crit. Rev. Anal. Chem.} \textbf{2025}, 1--43.

\bibitem{Abdullakutty2024}
Abdullakutty, F.; Akbari, Y.; Al-Maadeed, S.; Bouridane, A.; Talaat, I.M.; Hamoudi, R. Histopathology in focus: A review on explainable multi-modal approaches for breast cancer diagnosis. \textit{Front. Med.} \textbf{2024}, \textit{11}, 1450103.

\bibitem{Stefanelli2019}
Stefanelli, M. Applications of Expert Systems Technology to Medicine. In \textit{Artificial Intelligence}; Routledge: London, UK, 2019; pp. 233--264.

\bibitem{Tiwari2025}
Tiwari, A.; Mishra, S.; Kuo, T.R. Current AI technologies in cancer diagnostics and treatment. \textit{Mol. Cancer} \textbf{2025}, \textit{24}, 159.

\bibitem{Li2025Pathological}
Li, W.; Zhang, Y.; Yang, S.; Zhang, X.; Li, P.; Wang, R. Pathological graph self-supervised learning for clear-cell renal cell carcinoma survival prediction. \textit{Pattern Recognit.} \textbf{2025}, \textit{160}, 112572.

\bibitem{Chan2023}
Chan, T.H.; Cendra, F.J.; Ma, L.; Yin, G.; Yu, L. Histopathology whole slide image analysis with heterogeneous graph representation learning. In Proceedings of the IEEE/CVF Conference on Computer Vision and Pattern Recognition (CVPR), Vancouver, BC, Canada, 18--22 June 2023; pp. 15661--15670.

\bibitem{Sun2023}
Sun, C.; Li, C.; Lin, X.; Zheng, T.; Meng, F.; Rui, X.; Wang, Z. Attention-based graph neural networks: A survey. \textit{Artif. Intell. Rev.} \textbf{2023}, \textit{56}, 2263--2310.

\bibitem{Yin2024Hierarchical}
Yin, F.; Lam, H.K.; Watson, D. Hierarchical fuzzy model-agnostic explanation: Framework, algorithms and interface for XAI. \textit{IEEE Trans. Fuzzy Syst.} \textbf{2024}, \textit{32}, 2781--2795.

\bibitem{Cang2018Representability}
Cang, Z.; Mu, L.; Wei, G.W. Representability of algebraic topology for biomolecules in machine learning based scoring and virtual screening. \textit{PLoS Comput. Biol.} \textbf{2018}, \textit{14}, e1005929.

\bibitem{Pei2024Credibility}
Pei, Z.; Deng, L.; Xu, Y.; Li, M.; Xu, L.; Yan, L. Credibility of a membership function related to a linguistic value to improve computing with words. \textit{IEEE Trans. Fuzzy Syst.} \textbf{2024}, \textit{32}, 3458--3470.

\bibitem{Muhammad2025ALL}
Muhammad, D.; Salman, M.; Keles, A.; Bendechache, M. ALL diagnosis: Can efficiency and transparency coexist? An explainable deep learning approach. \textit{Sci. Rep.} \textbf{2025}, \textit{15}, 12812.

\bibitem{b6}
Spanhol, F.A.; Oliveira, L.S.; Petitjean, C.; Heutte, L. A dataset for breast cancer histopathological image classification. \textit{IEEE Trans. Biomed. Eng.} \textbf{2015}, \textit{63}, 1455--1462.

\bibitem{b7}
Lekamlage, C.D.; Afzal, F.; Westerberg, E.; Cheddad, A. Mini-DDSM: Mammography-based automatic age estimation. In Proceedings of the 3rd International Conference on Digital Medicine and Image Processing, Kyoto, Japan, 6--9 November 2020; pp. 1--6.

\bibitem{b8}
Aresta, G.; Araújo, T.; Kwok, S.; Chennamsetty, S.S.; Safwan, M.; et al. BACH: Grand challenge on breast cancer histology images. \textit{Med. Image Anal.} \textbf{2019}, \textit{56}, 122--139.

\bibitem{He2016}
He, K.; Zhang, X.; Ren, S.; Sun, J. Deep residual learning for image recognition. In Proceedings of the IEEE Conference on Computer Vision and Pattern Recognition (CVPR), Las Vegas, NV, USA, 27–30 June 2016; pp. 770--778.

\bibitem{Kipf2016}
Kipf, T.N.; Welling, M. Semi-supervised classification with graph convolutional networks. \textit{arXiv} \textbf{2016}, arXiv:1609.02907.

\bibitem{Dosovitskiy2020}
Dosovitskiy, A.; Beyer, L.; Kolesnikov, A.; Weissenborn, D.; Zhai, X.; et al. An image is worth 16x16 words: Transformers for image recognition at scale. \textit{arXiv} \textbf{2020}, arXiv:2010.11929.

\bibitem{Liu2021}
Liu, Z.; Lin, Y.; Cao, Y.; Hu, H.; Wei, Y.; Zhang, Z.; Lin, S.; Guo, B. Swin transformer: Hierarchical vision transformer using shifted windows. In Proceedings of the IEEE/CVF International Conference on Computer Vision (ICCV), Montreal, QC, Canada, 11--17 October 2021; pp. 10012--10022.

\bibitem{Dai2021}
Dai, Z.; Liu, H.; Le, Q.V.; Tan, M. CoAtNet: Marrying convolution and attention for all data sizes. \textit{Adv. Neural Inf. Process. Syst.} \textbf{2021}, \textit{34}, 3965--3977.

\bibitem{Liu2022}
Liu, Z.; Mao, H.; Wu, C.Y.; Feichtenhofer, C.; Darrell, T.; Xie, S. A convnet for the 2020s. In Proceedings of the IEEE/CVF Conference on Computer Vision and Pattern Recognition (CVPR), New Orleans, LA, USA, 18--24 June 2022; pp. 11976--11986.

\bibitem{Velickovic2017}
Veličković, P.; Cucurull, G.; Casanova, A.; Romero, A.; Liò, P.; Bengio, Y. Graph attention networks. \textit{arXiv} \textbf{2017}, arXiv:1710.10903.

\bibitem{Adnan2025}
Adnan, T.; Abdelkader, A.; Liu, Z.; Hossain, E.; Park, S.; Islam, M.S.; Hoque, E. A novel fusion architecture for detecting Parkinson’s Disease using semi-supervised speech embeddings. \textit{npj Parkinsons Dis.} \textbf{2025}, \textit{11}, 176.

\bibitem{Zhao2025}
Zhao, Z.; Yan, X.; Xiang, L.; Jiang, D. Fault Diagnosis of Rolling Bearings in High-Noise Environments Based on a Variational Gated Hierarchical Memory Network. \textit{Meas. Sci. Technol.} \textbf{2025}, doi:10.1088/1361-6501/ad9c8f.

\bibitem{Choudhry2025}
Arshad Choudhry, I.; Iqbal, S.; Alhussein, M.; Aurangzeb, K.; Qureshi, A.N.; Hussain, A. A novel interpretable graph convolutional neural network for multimodal brain tumor segmentation. \textit{Cogn. Comput.} \textbf{2025}, \textit{17}, 24.

\bibitem{b9}
Carrington, A.M.; Manuel, D.G.; Fieguth, P.W.; Ramsay, T.; Osmani, V.; et al. Deep ROC analysis and AUC as balanced average accuracy, for improved classifier selection, audit and explanation. \textit{IEEE Trans. Pattern Anal. Mach. Intell.} \textbf{2022}, \textit{45}, 329--341.

\bibitem{b10}
Chicco, D.; Tötsch, N.; Jurman, G. The Matthews correlation coefficient (MCC) is more reliable than balanced accuracy, bookmaker informedness, and markedness in two-class confusion matrix evaluation. \textit{BioData Min.} \textbf{2021}, \textit{14}, 1--22.

\bibitem{b11}
Parekh, D.H.; Dahiya, V. Predicting breast cancer using machine learning classifiers and enhancing the output by combining the predictions to generate optimal F1-score. \textit{Biomed. Biotechnol. Res. J.} \textbf{2021}, \textit{5}, 331--334.

\bibitem{b12}
Wetstein, S.C.; de Jong, V.M.; Stathonikos, N.; Opdam, M.; Dackus, G.M.; et al. Deep learning-based breast cancer grading and survival analysis on whole-slide histopathology images. \textit{Sci. Rep.} \textbf{2022}, \textit{12}, 15102.



\bibitem{DeBock1994}
De Bock, G.H.; Houwing-Duistermaat, J.J.; Springer, M.P.; Kievit, J.; Van Houwelingen, J.C. Sensitivity and specificity of diagnostic tests in acute maxillary sinusitis determined by maximum likelihood in the absence of an external standard. \emph{Journal of Clinical Epidemiology} \textbf{1994}, \emph{47}(12), 1343--1352.


\bibitem{Betti2009Role}
Betti, I.; Castelli, G.; Barchielli, A.; Beligni, C.; Boscherini, V.; De Luca, L.; et al. The role of N-terminal PRO-brain natriuretic peptide and echocardiography for screening asymptomatic left ventricular dysfunction in a population at high risk for heart failure. The PROBE-HF study. \emph{Journal of Cardiac Failure} \textbf{2009}, \emph{15}(5), 377--384.

\bibitem{Dad2025}
Dad, I.; He, J.; Baloch, Z. Graph-based analysis of histopathological images for lung cancer classification using GLCM features and enhanced graph. \textit{Front. Oncol.} \textbf{2025}, \textit{15}, 1546635.

\bibitem{Xiao2025Artificial}
Xiao, Z.; Feng, B.; Yang, J.; Sun, G.; Shen, Y.; Xu, S.; et al. Artificial intelligence in pathology: Advancing large models for scalable applications. \textit{Annu. Rev. Biomed. Data Sci.} \textbf{2025}, \textit{8}, 149--171.

\bibitem{Zhang2025Content}
Zhang, Y.; Zhang, X.; Qi, X.; Wu, X.; Chen, F.; Yang, G.; Fu, H. Content generation models in computational pathology: A comprehensive survey on methods, applications, and challenges. \textit{IEEE Rev. Biomed. Eng.} \textbf{2025}, \textit{18}, 1--18.

\bibitem{Yu2025ClinSegNet}
Yu, B.; Markham, H.; Moutasim, K.; Foria, V.; Liu, H. ClinSegNet: Towards reliable and enhanced histopathology screening. \textit{Bioengineering} \textbf{2025}, \textit{12}, 1156.

\bibitem{Cheng2025}
Cheng, X.; Ding, W.; Huang, J.; Ju, H.; Zhou, T.; Guo, J.; Pedrycz, W. FCAformer: Fuzzy-Enhanced Class-Aware Attention Based Transformer for Weakly Supervised Histopathology Image Segmentation. \textit{IEEE Trans. Fuzzy Syst.} \textbf{2025}, doi:10.1109/TFUZZ.2024.3510526.

\bibitem{Zhang2025}
Zhang, Z.; Zhou, Y.; Qi, Y.; Zhu, X.; Deng, X.; Tan, F.; Hu, P. Leveraging 3D Molecular Spatial Visual Information and Multi‐Perspective Representations for Drug Discovery. \textit{Adv. Sci.} \textbf{2025}, doi:10.1002/advs.202412453.

\bibitem{Cao2023}
Cao, L.; Wang, J.; Zhang, Y.; Rong, Z.; Wang, M.; Wang, L.; Hou, Y. E2EFP-MIL: End-to-end and high-generalizability weakly supervised deep convolutional network for lung cancer classification from whole slide image. \textit{Med. Image Anal.} \textbf{2023}, \textit{88}, 102837.

\end{thebibliography}
\end{document}